\documentclass{article} 
\usepackage{iclr2026_conference,times}
\iclrfinalcopy

\usepackage{amsmath,amsfonts,bm}

\def\eqref#1{equation~\ref{#1}}

\def\1{\bm{1}}

\DeclareMathAlphabet{\mathsfit}{\encodingdefault}{\sfdefault}{m}{sl}
\SetMathAlphabet{\mathsfit}{bold}{\encodingdefault}{\sfdefault}{bx}{n}

\usepackage{url}

\usepackage[utf8]{inputenc} 
\usepackage[T1]{fontenc}    
\usepackage{url}            
\usepackage{booktabs}       
\usepackage{amsfonts}       
\usepackage{nicefrac}       
\usepackage{microtype}      
\usepackage{xcolor}         
\usepackage{natbib}
\usepackage[utf8]{inputenc} 
\usepackage[T1]{fontenc}    
\usepackage{url}            
\usepackage{booktabs}       
\usepackage{amsfonts}       
\usepackage{nicefrac}       
\usepackage{microtype}      
\usepackage[table]{xcolor}         
\usepackage{graphicx}
\usepackage{caption}
\usepackage{booktabs}
\usepackage{amsmath}
\usepackage{amssymb}
\usepackage{pifont}
\usepackage{subcaption}
\usepackage{stfloats}
\usepackage{wrapfig}
\usepackage{multirow}
\usepackage{amssymb}
\usepackage{xcolor}
\usepackage{xspace}
\usepackage{dsfont}
\usepackage{enumitem}
\usepackage{wrapfig}  
\usepackage{xcolor}
\definecolor{HyperlinkBlue}{RGB}{0, 102, 204} 
\definecolor{MyLinkColor}{HTML}{092997}  
\definecolor{darkred}{RGB}{156, 39, 33}
\definecolor{darkblue}{RGB}{31, 90, 153}
\definecolor{mygray}{gray}{0.9}
\definecolor{myblue}{RGB}{174,199,232}
\definecolor{myblack}{RGB}{0,0,0}
\definecolor{citeblue}{HTML}{31ACD0}
\usepackage{arydshln}
\usepackage[
    colorlinks=true,
    linkcolor=HyperlinkBlue,
    urlcolor=HyperlinkBlue,
    citecolor=HyperlinkBlue,
    pdfborder={0 0 0}
]{hyperref}
\usepackage{tcolorbox}
\tcbuselibrary{skins, breakable, theorems}

\usepackage{listings}
\usepackage{array}
\usepackage{scalerel}
\usepackage{soul}
\usepackage{color}
\usepackage{longtable}
\newcommand{\git}{\raisebox{-1.5pt}{\includegraphics[height=1.05em]{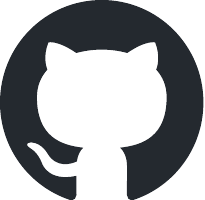}}\xspace}
\newcommand{\hf}{\raisebox{-1.5pt}{\includegraphics[height=1.05em]{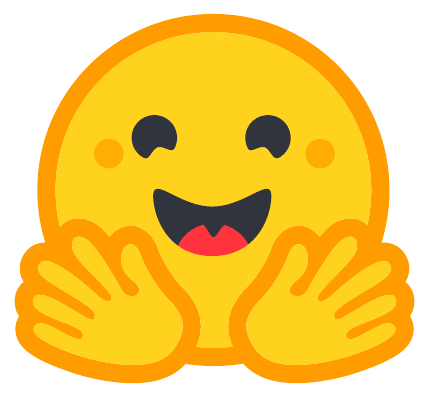}}\xspace}

\title{ \includegraphics[scale=0.04]{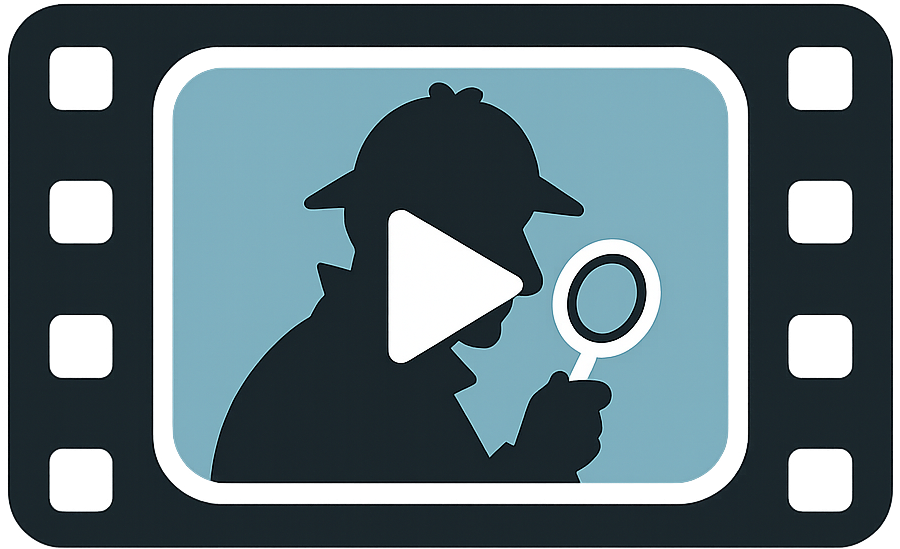} MMR-V: \textit{What's Left Unsaid?} A Benchmark for Multimodal Deep Reasoning in Videos}

\author{Kejian Zhu\textsuperscript{1,2}, Zhuoran Jin\textsuperscript{1,2}\footnotemark[2], Hongbang Yuan\textsuperscript{1,2}, Jiachun Li\textsuperscript{1,2}, Shangqing Tu\textsuperscript{3} \\ 
\textbf{Pengfei Cao\textsuperscript{1,2},}  \textbf{Yubo Chen\textsuperscript{1,2},}  \textbf{Kang Liu\textsuperscript{1,2},} \textbf{Jun Zhao\textsuperscript{1,2}\footnotemark[2]} \\ \textsuperscript{1}The Key Laboratory of Cognition and Decision Intelligence for Complex Systems,\\
 Institute of Automation, Chinese Academy of Sciences, Beijing, China \\ \textsuperscript{2}School of Artificial Intelligence, University of Chinese Academy of Sciences \textsuperscript{3}Tsinghua University\\
\texttt{zhukejian2025@ia.ac.cn } \texttt{\{zhuoran.jin, hongbang.yuan\} @nlpr.ia.ac.cn } \\
\texttt{\{pengfei.cao, yubo.chen, kliu, jzhao\} @nlpr.ia.ac.cn } 
}

\begin{document}

\maketitle
\renewcommand{\thefootnote}{\fnsymbol{footnote}}
\footnotetext[2]{Corresponding authors.}
\renewcommand*{\thefootnote}{\arabic{footnote}}
\begin{abstract}
The sequential structure of videos poses a challenge to the ability of multimodal large language models (MLLMs) to locate multi-frame evidence and conduct multimodal reasoning. However, existing video benchmarks mainly focus on understanding tasks, which only require models to match frames mentioned in the question (hereafter referred to as ``question frame'') and perceive a few adjacent frames. To address this gap, we propose \textbf{MMR-V: A Benchmark for Multimodal Deep Reasoning in Videos}. The benchmark is characterized by the following features. \textbf{(1) Long-range, multi-frame reasoning}: Models are required to infer and analyze evidence frames that may be far from the question frame. \textbf{(2) Beyond perception}: Questions cannot be answered through direct perception alone but require reasoning over hidden information. \textbf{(3) Reliability}: All tasks are manually annotated, referencing extensive real-world user understanding to align with common perceptions. \textbf{(4) Confusability}: Carefully designed distractor annotation strategies to reduce model shortcuts. MMR-V consists of 317 videos and 1,257 tasks. Our experiments reveal that current models still struggle with multi-modal reasoning; even the best-performing model, Gemini-2.5-pro, achieves only 64.3\% accuracy. Additionally, current reasoning enhancement strategies (Chain-of-Thought and scaling test-time compute) bring limited gains. Error analysis indicates that the CoT demanded for multi-modal reasoning differs from it in textual reasoning, which partly explains the limited performance gains. We hope that MMR-V can inspire further research into enhancing multi-modal reasoning capabilities.
\end{abstract}

{\footnotesize 
\urlstyle{rm}
\begin{center}
    \renewcommand{\arraystretch}{1.2}
    \vspace{-10pt}
    \begin{tabular}{rcl}
         \hf & \textbf{Benchmark} & \url{https://huggingface.co/datasets/JokerJan/MMR-VBench}\\
          \git & \textbf{Project} & \url{https://mmr-v.github.io/}        
    \end{tabular}
\end{center}
}

\section{Introduction}
\label{sec:introduction}

Recent models like OpenAI's o1~\citep{jaech2024openai} and Deepseek-R1~\citep{guo2025deepseek} have significantly improved text reasoning ability through reinforcement learning. This has sparked growing interest in multimodal reasoning~\citep{wang2025multimodal}. Models like o3~\citep{o3} and GPT-5~\citep{gpt5} have achieved impressive results on image reasoning tasks through tool use, integrating visual information into the reasoning process to enable deep reflection and evidence mining. However, most of these studies focus on images, with limited exploration of more challenging video reasoning tasks. Video involves sequential and richer multimodal information, requiring models to reason and mine evidence over long-range, multi-frame. Since this capability is essential for real-world applications such as embodied intelligence and intelligent security monitoring~\citep{hou2008research, yang2024thinking}, it naturally raises an important question: \textit{can current MLLMs perform deep multimodal reasoning and ``think with videos'' like o3 on image tasks?}

\begin{figure}[t]
\vspace{-15pt}
    \centering
    \includegraphics[width=\linewidth]{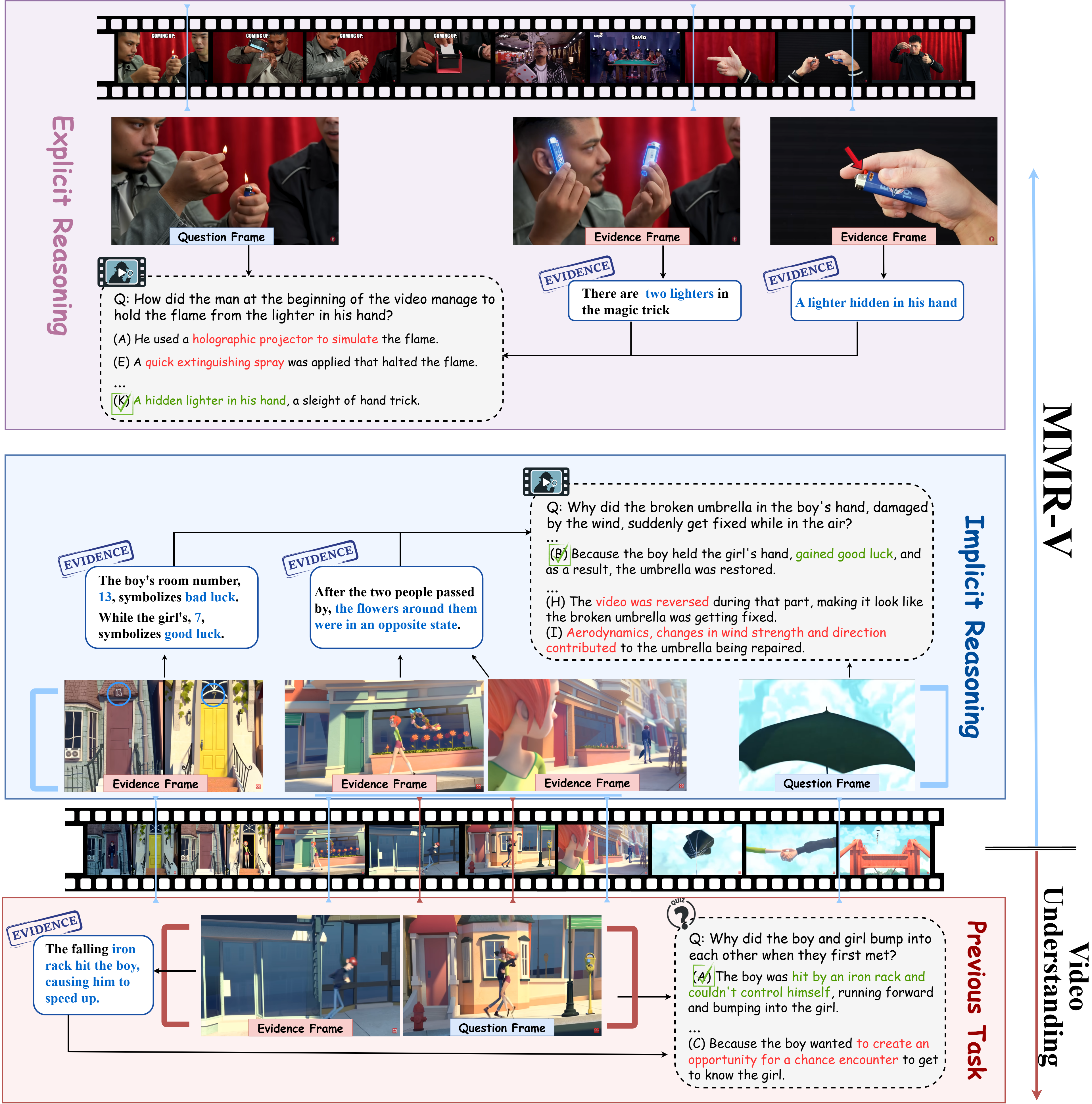}
    \caption{Examples showing the MMR-V tasks and the difference from previous video benchmarks.}
    \label{fig:benchmark_intro}
\vspace{-10pt}
\vspace{-1em}
\end{figure}

However, existing video benchmarks primarily focus on perception and understanding tasks~\citep{zhou2024mlvu, fu2024video}. These tasks often only require locating frames mentioned in the question and understanding adjacent frames. For example, at the bottom of Figure \ref{fig:benchmark_intro}, noticing the boy being hit by the metal frame is enough to understand why he ran into the girl. Such tasks fall short in evaluating multimodal reasoning abilities. We summarize their limitations as follows:
\textbf{(1) Limited frame context}: 
Even for long videos, existing tasks often rely on just a few adjacent frames, failing to exploit the long-range sequential structure of the video.
\textbf{(2) Lack of reasoning}: Many questions can be answered through direct perception.  
\textbf{(3) Unrealistic task}: Simple perception and adjacent-frame understanding tasks do not meet the real-world demands for AI system strong capabilities.

To address these shortcomings, we propose \textbf{MMR-V Bench: A Benchmark for Multi-modal Deep Reasoning in Videos}. We present two examples to illustrate the key differences with previous video understanding benchmarks in Figure~\ref{fig:benchmark_intro}. MMR-V offers the following features: \textbf{(1) Long-range, multi-frame reasoning:} tasks involve multimodal reasoning over non-adjacent video frames to locate and analyze multiple evidences;  \textbf{(2) Beyond perception: }questions cannot be answered by direct perception of question frame directly, requiring reasoning and the extraction of implications; \textbf{(3) Reliability: }All tasks are annotated manually, and potential subjective bias is reduced by cross-referencing the most popular video comments. \textbf{(4) Confusability:} We employ carefully designed annotation strategies to craft model-aligned distractor options, thereby ensuring confusability.


Inspired by cognitive and psychological theories~\citep{evans1984heuristic, sun2006clarion, polanyi2012personal}, such as Kahneman's Dual Process Theory~\citep{kahneman2011thinking}, we categorize the tasks in MMR-V into \textbf{implicit reasoning} and \textbf{explicit reasoning}. The key distinction lies in \textit{whether the question requires reasoning beyond surface-level information to infer underlying implications}. Explicit reasoning is defined as questions that can be solved using perceivable information from the video. For example, the task shown in Figure \ref{fig:benchmark_intro} requires noticing the two lighters hidden in the hand. Implicit reasoning requires extracting and interpreting the underlying subtext behind visual information. For example, in the implicit reasoning case shown in Figure \ref{fig:benchmark_intro}, it requires inferring the underlying implication that the girl’s room number 7 symbolizes good luck. This is more of an assessment of \textit{EQ}, testing whether the model can use its deep understanding of the world knowledge to make implicit and subconscious reasoning paths like humans. 


\textbf{MMR-V comprises 317 videos and 1257 tasks}. The videos span six major categories, with lengths ranging from 7 to 3771 seconds, with an average of 277 seconds. Tasks are further divided into 10 categories and subcategories. Each task is in multiple-choice format with approximately ten options on average. Tasks typically require reasoning over average 12 video frames, covering about 60\% of video duration. All questions and correct answers are human-annotated and reviewed. Distractors are generated using a carefully designed annotation strategy (Details in Section \ref{sec:Data Annotation}). 

We evaluated 11 proprietary models and 10 open-source models on MMR-V. The results reveal that even the best-performing model, Gemini-2.5-pro, achieved \textbf{only 64.3\% accuracy}, highlighting the significant challenge MMR-V poses to current multimodal large language models. Our key findings are as follows. \textbf{(1) Multimodal reasoning challenge:} 
Our findings in Section~\ref{exp:main results} show that reasoning enhancement strategies (e.g., CoT and scaling test-time compute) yield limited improvements, indicating that MMR-V presents a greater challenge to current multimodal reasoning models. Further error analysis in Section~\ref{sec:error analysis} shows that the CoT demanded in multimodal reasoning differs from those in textual reasoning. Current models tend to rely on textual reasoning based on visual information from the question frame and few adjacent frames, lacking the multimodal reasoning needed to locate and analyze evidence from long-range frames. This limitation hinders the overall reasoning performance. \textbf{(2) More modality will benefit:} We found that for models that support all modalities, adding additional audio modalities will improve the performance (Accuracy improved by 1.4\%, 1.0\%, and 1.0\% for Gemini 2.0-Flash, Gemini 2.0-Flash-Thinking, and Phi-4-Multimodal-Instruct, respectively). \textbf{(3) Human-model gap:} In human experiments, we found that although models exhibit human-level performance on text reasoning tasks, there is still a significant gap between model and human on multimodal, especially video, reasoning tasks. We hope MMR-V will inspire further research into enhancing multimodal reasoning capabilities in AI systems.


\section{Task Overview}

The tasks in MMR-V require deeper multimodal reasoning. Unlike previous tasks such as math and puzzle problems~\citep{lu2023mathvista, wang2024measuring, zhang2024humaneval}, we argue that the scope of multimodal reasoning should be more broadly defined. Previous work focuses more on text-oriented reasoning based on perceived visual information. In contrast, our task requires integrating the various forms of visual evidences, such as artistic style, lighting, and depth, into the reasoning process. Even more challenging, it involves \textbf{reasoning over long-range, multi-frame visual evidence}. Videos have a temporal dimension, which puts a greater challenge on the ability to find clues in different frames through multimodal reasoning. 

\subsection{Definition for Implicit and Explicit Reasoning.} 

\begin{figure}[t]
\vspace{-20pt}
    \centering
    \includegraphics[width=\linewidth]{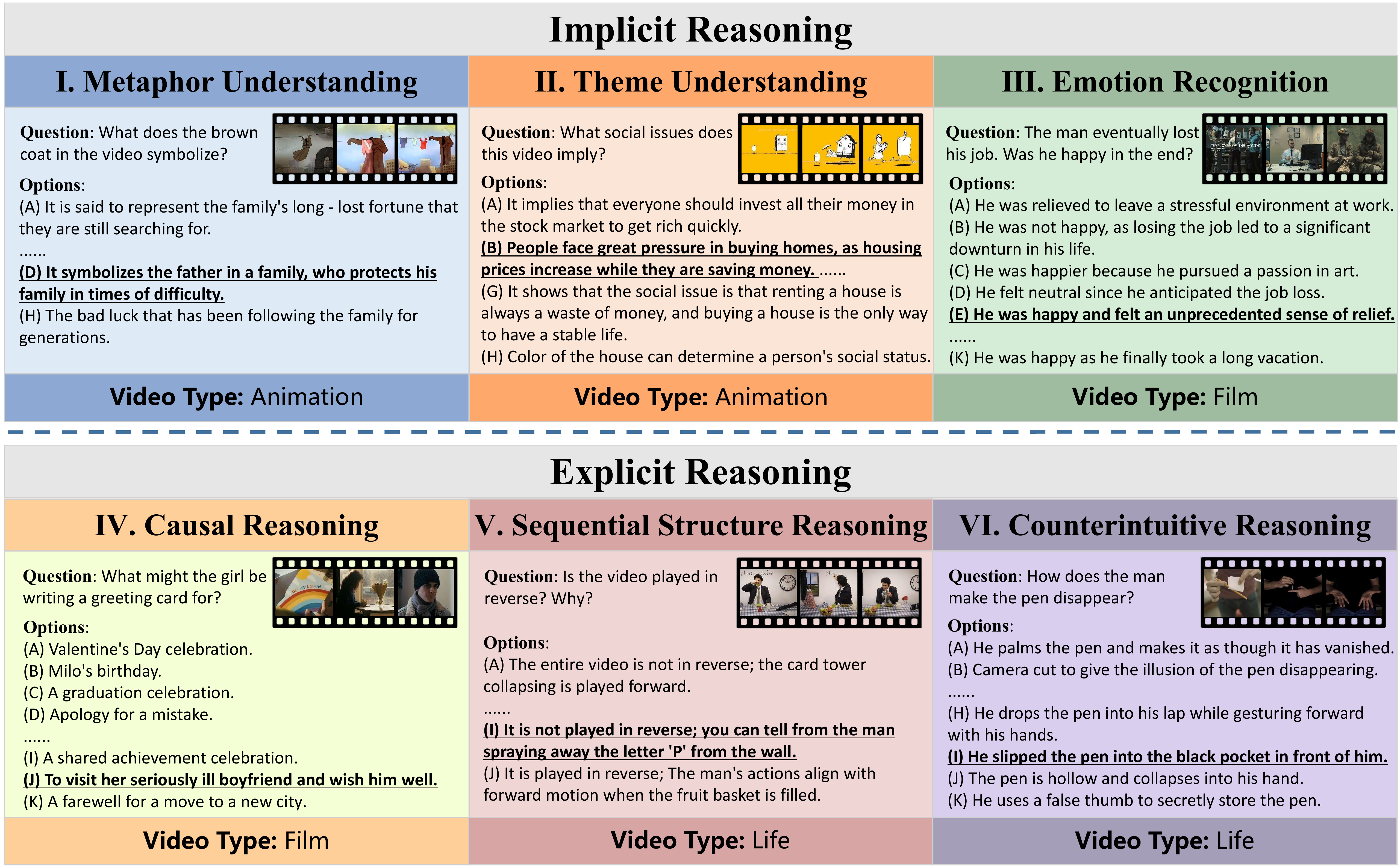}
    \caption{Overview of six tasks in MMR-V Bench.}
    \label{fig:task_overview}
\vspace{-15pt}
\end{figure}

We categorize reasoning tasks in MMR-V into \textbf{Implicit Reasoning} and \textbf{Explicit Reasoning}, inspired by Kahneman's Dual Process Theory~\citep{kahneman2011thinking} and other cognitive theories~\citep{evans1984heuristic, sun2006clarion, polanyi2012personal}. The most obvious difference is whether or not one needs to understand the subtext beneath the surface information. Secondly, implicit reasoning for human is often achieved by experience based on world knowledge, thus consuming little attention resources. Tasks are further divided into 10 categories and 33 subcategories. Figure \ref{fig:task_overview} shows six categories (top row: implicit; bottom: explicit). Further explanations and examples can be found in Appendix~\ref{sec:task details}.


\textbf{Implicit Reasoning} focuses on incorporating \textbf{hidden meanings behind visual information} into reasoning. In these tasks, surface-level visual cues often conceal deeper layers of meaning, such as metaphor. Besides, for human, \textit{“(implicit) operates automatically and quickly, with little or no effort and no sense of voluntary control.” - Dual Process Theory.}

\textbf{Explicit Reasoning} evaluates whether a model can perform reasoning based on multimodal details \textbf{explicitly presented} across long-range, multi-frame of a video. However, solving these tasks demands fine-grained perception and rigorous logical reasoning. \textit{“(explicit) allocates attention to the effortful mental activities that demand it, including complex computations.” - Dual Process Theory.}


\subsection{Implicit Reasoning Tasks}

\textbf{Metaphor Understanding (MU)}: MU tasks evaluate the ability to reason about metaphors for entities or environment. For example, the case in Figure \ref{fig:task_overview} I interprets the metaphor of the brown coat.

\textbf{Theme Understanding (TU)}: TU assesses the ability to infer the main idea and attitude of the author through the full video. For example, the case in Figure \ref{fig:task_overview} II asks what social issue the video reveals.

\textbf{Emotion Recognition (ER)}: ER tasks evaluate the ability to analyze character emotional states, as well as higher-level emotions such as the author’s attitude and the audience’s emotional response. For example, the case in Figure \ref{fig:task_overview} III involves inferring whether the character feels happy at the end.

\textbf{Comment Matching (CM)}: CM task is to predict the most fitting audience comments for a video based on a criteria. For example, selecting which comment would be the most humorous after watching the video. Detailed example can be found in Appendix \ref{Appendix: CM}.

\textbf{Implicit Symbol (IS)}: IS task is to interpret implicit symbols in the video, such as cultural elements. For example, inferring the ethnicity of the filming location. Details can be found in Appendix \ref{Appendix: IS}.

\subsection{Explicit Reasoning Tasks}

\textbf{Causal Reasoning (CAR)}: CAR assesses the ability to reason about causal relationships in the video. For example, in Figure \ref{fig:task_overview} IV, it involves inferring the reason why the girl is making a card.

\textbf{Sequential Structure Reasoning (SSR)}: SSR tasks assess reasoning about temporal structure in video editing and storytelling. In the example from Figure \ref{fig:task_overview} V, the task is to infer if the video is reversed. However, the creator of this video explains the video is played normally.

\textbf{Counterintuitive Reasoning (CIR)}: CIR tasks evaluate the ability to analyze information that contradicts common sense, requiring detailed cross-frame analysis. In the example from Figure \ref{fig:task_overview} VI, the task is to reason the principle behind the counterintuitive magic trick.

\textbf{Cross-modal Transfer Reasoning (CTR)}: To reason and match information out of the video that shares similar meaning. For example, find the quote with same theme of the video.

\textbf{Video Type and Intent (VTI)}: VTI tasks test the ability to infer key meta-level information such as the genre and communicative intent of the video from a global perspective. For example, the case in Appendix \ref{Appendix: VTI} infers the release time by reasoning the video is set during COVID-19.

\section{MMR-V Bench}

To ensure that MMR-V effectively evaluates multimodal reasoning abilities, we follow \textbf{three principles} during construction:  
\textbf{P1. Multi-frame:} Questions require reference to long-range, multi-frame information, prompting the model to reason across multiple visual cues.  
\textbf{P2. Deep reasoning:} Answers should not be directly perceivable from the video; instead, they should demand understanding of the subtext or multimodal reasoning, reflecting a deep comprehension of the content.  
\textbf{P3. Realistic:} Tasks should align with real-world question-answering needs, ensuring answers are consistent with common user understanding and free from individual cognitive biases or prejudices.

\label{principles}

\subsection{Video Collection}
\label{Video Collection}

We manually curated diverse original videos from Youtube with the following checklist: (1)\textbf{ Avoidance of linear, descriptive content}: We excluded videos with straightforward structures, such as daily recordings or sports broadcasts, in order to ensure that the tasks require deep reasoning over multi-frames (For Principle P1). (2)\textbf{ Creative and thematically rich videos}: We selected videos that are intentional designed and edited by creators, often conveying well-crafted themes. This ensures that the questions require interpretation beyond surface-level visual content (For Principle P2). (3) \textbf{Alignment with real-world}: Highly Popular Videos were preferred, which are indicated by active comment sections and audience engagement. This helps avoid biases introduced by niche content and ensures alignment with general user cognition (For Principle P3). (4)\textbf{ Diverse coverage}: To further promote generalizability, we ensured broad coverage across video types, topics, and durations, allowing MMR-V to reflect the diversity of real-world video content (For Principle P3). As a result, our final benchmark comprises \textbf{317 videos} spanning \textbf{six major categories}: Animation, Film, Philosophy, TV, Life, and Art. The specific categories are shown in the Appendix \ref{sec:diversity}. Furthermore, for problems where audio might be helpful, we ensure that the videos include audio.

\subsection{Data Annotation \& Quality Assurance}
\label{sec:Data Annotation}
All tasks in MMR-V Bench are designed in a multiple-choice format. There is one correct option and several wrong options. We make sure there are carefully crafted distractors among the wrong options. To ensure the quality and plausibility of these distractors, we designed three distinct distractors annotation strategies. (1) Str. 1: We prompt a strong model GPT-4o ~\citep{hurst2024gpt} to directly answer the manually annotated question. If the model generates an incorrect answer (as verified by human annotators), that answer is retained as a high-quality distractor. If correct, we combine human-written distractors with incorrect options generated by GPT-4o as distractors. (2) Str. 2: Given the question and correct answer annotated manually, GPT-4o is prompted to generate distractors. (3) Str. 3: Human annotators construct distractors manually.
 
\begin{wraptable}{r}{0.48\textwidth}  
\vspace{-12pt}
  \centering
  \fontsize{10pt}{12pt}\selectfont
  \caption{\label{exp:Annotation Strategies}
  Performance on 100 questions annotated with different strategies (str.).}
  \begin{tabular}{l|c|c|c}
    \toprule
    \textbf{Models}       & \textbf{Str. 1} & \textbf{Str. 2} & \textbf{Str. 3} \\
    \midrule
    GPT-4o                & \cellcolor[HTML]{FED9C7}59\% & 70\%  & 62\% \\
    Qwen-VL-7B           & \cellcolor[HTML]{FED9C7}37\% & 51\%  & 42\% \\
    \bottomrule
  \end{tabular}
\vspace{-5pt}
\end{wraptable}

We conducted a test using 100 questions, using three strategies to form three test-set with 100 multiple-choice tasks. As shown in Table \ref{exp:Annotation Strategies}, distractors generated by strategy 1 are more confusing, significantly increasing the difficulty and quality of our tasks.
It is worth noting that in the above test process, when GPT-4o directly answered 100 tasks, the accuracy rate verified by humans was only 17\%. This reflects the limitations of the current model in multimodal reasoning capabilities.

To ensure high quality, we also developed an checklist based on the construction principles and invited human annotators to verify the accuracy and difficulty of the tasks using this checklist. We invited five annotators with at least a bachelor's degree to participate in the annotation and review process. The checklist of MMR-V is shown in the Appendix \ref{sec:construction}. The overall annotation process and the annotation platform can be found in Figure \ref{fig:pipeline} and Figure \ref{fig:platform} in the Appendix \ref{sec:construction}.

\vspace{-5pt}
\subsection{Data Statistics}
\label{sec:statistics}
\begin{wraptable}{r}{0.4\textwidth} 
\vspace{-30pt}
  \caption{\label{table: Data Statistic}
  Dataset Statistic of MMR-V.}
  \centering
  \setlength\tabcolsep{6pt}
  \fontsize{10pt}{10pt}\selectfont
  \begin{tabular}{lc}
    \toprule
    \textbf{Dataset}       & \textbf{Statistic} \\
    \midrule
    \multicolumn{2}{l}{\emph{Task}} \\
    Question Count                & 1257 \\
    Average Option Count           & 10 \\
    Average Question Words           & 14 \\
    Average Option Words           & 10 \\
    \midrule
    \multicolumn{2}{l}{\emph{Video}} \\
    Video Count           & 317 \\
    Minimum Length (s)           & 7 \\
    Maximum Length (s)          & 3771 \\
    Average Length (s)          & 277 \\
    \bottomrule
  \end{tabular}

\end{wraptable}
MMR-V comprises a total of 317 videos spanning a wide range of content types, and includes 1,257 multiple-choice reasoning tasks. 
Each question is annotated with 7 to 11 candidate answers, with only one correct answer guaranteed. As illustrated in Figure \ref{fig:video-categories}, the videos are categorized into six major domains, each encompassing fine-grained subcategories to ensure diversity in content, style, and semantics. The reasoning tasks in our benchmark are organized across three levels of granularity, reflecting different dimensions of reasoning complexity and modality. The distribution of task types across these levels is shown in Figure \ref{fig:ability-type}. More information is shown in Table \ref{table: Data Statistic}. Table \ref{tab:benchmark_comparison} shows a statistic comparison with previous benchmarks.

\section{Experiments}
\label{Exp}

\subsection{Settings}
\label{sec:exp_setup}
We conducted extensive evaluations on 11 proprietary and 10 open-source models as detailed in the Appendix \ref{appendix:baselines}. Our main experiments were conducted under two settings: zero-shot and zero-shot + CoT~\citep{wei2022chain}, in order to examine whether reasoning enhances performance. For further analysis, we introduced the following categories of comparative models: (1) Models with different scales. (2) ``Thinking'' model and its base version. (e.g., Gemini-Flash and Gemini-Flash-Thinking).

\textbf{Multimodal Inputs}: For models supporting full-modal inputs (e.g., Gemini-2.0-flash), we further compare their performance with and without audio input to evaluate its influence on reasoning.

\textbf{Frame Selection}: Since some models only support multiple images or short video clips, we standardized the number of input frames. Details of frame sampling are provided in Appendix \ref{sec: exp details}.

\textbf{Human Experiment:} To provide a meaningful upper bound for MMR-V and to examine the human-model gap, we invited participants with at least bachelor degree to conduct human experiment. We sampled 100 tasks GPT-4o answered incorrectly and 100 tasks it answered correctly for experiment. 

\subsection{Main Results}
\label{exp:main results}


Results in Table \ref{tab:main exp} highlight the challenge of MMR-V. The highest score was achieved by Gemini-2.5-pro with 1 fps video sampling, reaching only 64.3\%. Under the setting of fixed input frame number, GPT-5 obtained the best score of 60.9\%. Among open-source models, Gemma-3-27b-it performs the best, demonstrating relatively strong performance. However, there remains a gap compared to proprietary models. The performance on videos of different lengths is shown in the Appendix \ref{sec:appendix_video_length}.

\begin{table*}[t]
\vspace{-15pt}
\centering
\normalsize
\renewcommand{\arraystretch}{1.07}
\setlength{\tabcolsep}{3.8pt}
\caption{Evaluation results (\%) on MMR-V. Results under \colorbox{mygray}{CoT} prompting are highlighted in gray. The random accuracy on MMR-V Bench is approximately 10\%. \textbf{Bold} and \underline{underlined} values indicate the best performance among proprietary and open-source models, respectively.}
\resizebox{\linewidth}{!}{%
\begin{tabular}{p{4.5cm}|m{0.7cm}|m{0.7cm}|m{0.7cm}|m{0.7cm}|m{0.7cm}|m{0.7cm}|m{0.7cm}|m{0.7cm}|m{0.7cm}|m{0.7cm}|m{0.7cm}|m{0.7cm}}
\hline
 &  \multicolumn{2}{c}{} & \multicolumn{4}{|c|}{\textbf{Tasks}} & \multicolumn{6}{c}{\textbf{Video Categories}} \\
\cline{2-13}
\textbf{Model} & \multicolumn{2}{c}{\textbf{Overall}} & \multicolumn{2}{|c|}{\textbf{Implicit}} & \multicolumn{2}{c|}{\textbf{Explicit}} & \textbf{Art} & \textbf{Life} & \textbf{TV} & \textbf{Film} & \textbf{Ani.} & \textbf{Phi.} \\ 
\hline
\multicolumn{13}{c}{\emph{Open-source models}} \\
\textrm{LLaVA-Video} & 18.4 & \cellcolor{mygray}17.6 & 19.1 & \cellcolor{mygray}18.1 & 15.4 & \cellcolor{mygray}16.3 & 14.4 & 11.2 & 13.2 & 17.4 & 21.4 & 12.8 \\
\textrm{NVILA-8B-Video} & 25.5 & \cellcolor{mygray}25.3 & 26.2 & \cellcolor{mygray}24.2 & 23.9 & \cellcolor{mygray}25.9 & 17.3 & 21.3 & 23.5 & 21.6 & 38.0 & 21.8 \\ 
\textrm{Phi-4-multimodal-instruct} & 26.7 & \cellcolor{mygray}27.6 & 29.4 & \cellcolor{mygray}31.2 & 19.4 & \cellcolor{mygray}18.1 & 19.4 & 19.2 & 25.9 & 26.4 & 33.9 & 24.4 \\
\textrm{Cogvlm2-video-llama3} & 25.6 & \cellcolor{mygray}26.1 & 25.4 & \cellcolor{mygray}26.2 & 26.1 & \cellcolor{mygray}25.7 & 15.5 & 18.3 & 24.7 & 19.1 & 43.2 & 20.8 \\ 
\textrm{Qwen2.5-VL-7B} & 30.1 & \cellcolor{mygray}32.4 & 33.7 & \cellcolor{mygray}36.2 & 20.8 & \cellcolor{mygray}22.5 & 20.9 & 18.1 & 29.6 & 21.2 & 48.4 & 19.8 \\
\textrm{Intern3-8B} & 33.6 & \cellcolor{mygray}32.9 & 35.5 & \cellcolor{mygray}33.4 & 28.6 & \cellcolor{mygray}31.4 & 23.0 & 22.6 & 31.7 & 24.3 & 52.9 & 23.2 \\
\textrm{Gemma-3-12b-it} & 34.0 & \cellcolor{mygray}34.2 & 37.8 & \cellcolor{mygray}37.6 & 24.0 & \cellcolor{mygray}25.4 & 19.4 & 24.9 & 25.9 & 31.3 & 51.9 & 24.4 \\
\textrm{InternVL2.5-38B} & 39.9 & \cellcolor{mygray}39.7 & 43.8 & \cellcolor{mygray}43.7 & 29.9 & \cellcolor{mygray}29.4 & 30.4 & 28.8 & 30.4 & 37.2 & \underline{57.4} & 29.1 \\
\textrm{Qwen2.5-VL-72B} & 39.1 & \cellcolor{mygray}40.4 & 41.3 & \cellcolor{mygray}42.8 & \underline{33.4} & \cellcolor{mygray}\underline{34.3} & 28.9 & 28.2 & 29.1 & 36.5 & 55.6 & \underline{37.2} \\ 
\textrm{Gemma-3-27b-it} & \underline{42.0} & \cellcolor{mygray}\underline{41.1} & \underline{46.5} & \cellcolor{mygray}\underline{44.7} & 30.3 & \cellcolor{mygray}32.0 & \underline{31.7} & \underline{32.2} & \underline{35.5} & \underline{41.3} & 56.1 & 33.7 \\
\hline
\multicolumn{13}{c}{\emph{Proprietary models}} \\

\textrm{GPT-4o-mini-2024-07-18} & 34.8 & \cellcolor{mygray}35.2 & 38.0 & \cellcolor{mygray}38.6 & 26.3 & \cellcolor{mygray}26.3 & 29.5 & 25.4 & 29.6 & 33.0 & 48.7 & 18.6 \\
\textrm{Gemini-2.0-Flash (16 frames)} & 42.6 & \cellcolor{mygray}44.3 & 44.3 & \cellcolor{mygray}45.9 & 38.3 & \cellcolor{mygray}40.0 & 30.9 & 32.2 & 40.7 & 40.6 & 58.5 & 24.4 \\
\textrm{Claude-3.5-Sonnet-20241022} & 43.3 & \cellcolor{mygray}44.2 & 45.0 & \cellcolor{mygray}46.1 & 38.9 & \cellcolor{mygray}39.1 & 33.8 & 31.1 & 41.3 & 41.3 & 55.8 & 4.4\\
\textrm{Gemini-2.0-Flash-thinking} & 45.0 & \cellcolor{mygray}43.5 & 46.6 & \cellcolor{mygray}46.0 & 40.6 & \cellcolor{mygray}37.1 & 34.5 & 31.6 & 38.6 & 48.3 & 60.1 & 25.6 \\
\textrm{GPT-4.1-2025-04-14} & 46.6 & \cellcolor{mygray}48.9 & 49.1 & \cellcolor{mygray}51.7 & 40.3 & \cellcolor{mygray}41.7 & 43.2 & 35.6 & 43.9 & 46.5 & 57.1 & 34.9\\
\textrm{Gemini-2.0-Flash (512 frames)} & 48.0 & \cellcolor{mygray}49.9 & 50.5 & \cellcolor{mygray}52.6 & 41.6 & \cellcolor{mygray}42.9 & 36.7 & 36.7 & 39.7 & 46.2 & 66.7 & 31.4 \\
\textrm{Gemini-2.5-Flash} & 51.2 & \cellcolor{mygray}50.5 & 52.9 & \cellcolor{mygray}52.3 & 46.9 & \cellcolor{mygray}45.7 & 45.3 & 39.5 & 50.3 & 47.9 & 65.6 & 34.9 \\
\textrm{o4-mini-2025-04-16} & 52.5 & \cellcolor{mygray}52.1 & 54.6 & \cellcolor{mygray}54.5 & 47.1 & \cellcolor{mygray}46.0 & 48.2 & 40.1 & 54.0 & 51.7 & 65.3 & 27.9\\
\textrm{GPT-4o-2024-11-20} & 52.8  & \cellcolor{mygray}55.0 & 54.7 & \cellcolor{mygray}56.3 & 48.1 & \cellcolor{mygray}51.4 & 46.0 & 42.5 & 50.0 & 49.0 & 67.7 & 38.4\\
\textrm{o3-2025-04-16} & 59.1  & \cellcolor{mygray}58.3 & 60.1 & \cellcolor{mygray}57.8 & 57.6 & \cellcolor{mygray}59.7 & 51.8 & 45.2 & 55.6 & 58.3 & 74.3 & 43.0\\
\textrm{GPT-5-2025-08-07} & 60.6  & \cellcolor{mygray}60.9 & 61.0 & \cellcolor{mygray}60.6 & 59.7 & \cellcolor{mygray}\textbf{61.4} & 56.8 & 41.8 & 56.6 & 60.4 & \textbf{76.5}  & 45.4\\
\textrm{Gemini-2.5-pro (1fps)} & \textbf{64.3} & \cellcolor{mygray}\textbf{63.2} & \textbf{65.9} & \cellcolor{mygray}\textbf{64.4} & \textbf{60.2} & \cellcolor{mygray}60.3 & \textbf{57.6} & \textbf{58.1} & \textbf{65.6} & \textbf{62.5} & 73.0 & \textbf{54.6}\\
\hline
\multicolumn{13}{c}{\emph{Baseline}} \\
\cellcolor{myblue}\emph{Best Performance of Models} & \multicolumn{2}{c|}{\cellcolor{myblue}64.3} & \multicolumn{2}{c|}{\cellcolor{myblue}65.9} & \multicolumn{2}{c|}{\cellcolor{myblue}60.3} & \multicolumn{1}{c|}{\cellcolor{myblue}57.6} & \multicolumn{1}{c|}{\cellcolor{myblue}58.1} & \multicolumn{1}{c|}{\cellcolor{myblue}65.6} & \multicolumn{1}{c|}{\cellcolor{myblue}62.5} & \multicolumn{1}{c|}{\cellcolor{myblue}74.3} & \multicolumn{1}{c}{\cellcolor{myblue}54.6} \\
\cellcolor{myblue}\emph{Human} & \multicolumn{2}{c|}{\cellcolor{myblue}86.0} & \multicolumn{2}{c|}{\cellcolor{myblue}80.6} & \multicolumn{2}{c|}{\cellcolor{myblue}91.2} & \multicolumn{1}{c|}{\cellcolor{myblue}57.7} & \multicolumn{1}{c|}{\cellcolor{myblue}92.3} & \multicolumn{1}{c|}{\cellcolor{myblue}90.6} & \multicolumn{1}{c|}{\cellcolor{myblue}92.3} & \multicolumn{1}{c|}{\cellcolor{myblue}90.7} & \multicolumn{1}{c}{\cellcolor{myblue}70.0} \\
\hline
\end{tabular}%
}
\label{tab:main exp}
\vspace{-15pt}
\end{table*}

\textbf{Current reasoning enhancements have limitations on MMR-V. } 
Results in Table \ref{tab:main exp} show that current reasoning enhancement strategies, which are relatively effective in textual domains, such as CoT prompt reasoning and scaling test-time compute (i.e., "Thinking" models), offer only limited gains on MMR-V. CoT brings only a 0.88\% average gain, and "Thinking" model improves just 2.4\%. This indicates that MMR-V presents a significant challenge to the multimodal reasoning capabilities of existing models.
Analysis of sampled model responses shows that visual analysis accounts for only about 10\% of the CoTs. This reveals that reasoning process of current model is mostly text-based (reasoning on questions and options), relying on visual perception of question frame, instead of integrating visual reasoning and evidence mining into CoTs.
Several examples are provided in Appendix \ref{sec:case}, and further analysis in Section \ref{sec:error analysis} supports similar findings.

\textbf{Model performance on MMR-V Bench shows a clear scaling law effect.} Smaller models under the same architecture perform poorly on tasks that require complex reasoning. For instance, larger models like Qwen2.5-VL-72B (39.1\%) and GPT-4o (52.8\%) outperform their smaller versions Qwen2.5-VL-7B (30.1\%) and GPT-4o-mini (34.8\%), showing relative gains of 9\% and 18\%, respectively.

\textbf{Model performance across different tasks on MMR-V Bench.}

\label{sec:diff-tasks}
\begin{wrapfigure}{r}{0.45\textwidth}
  \centering
  \includegraphics[width=0.45\textwidth]{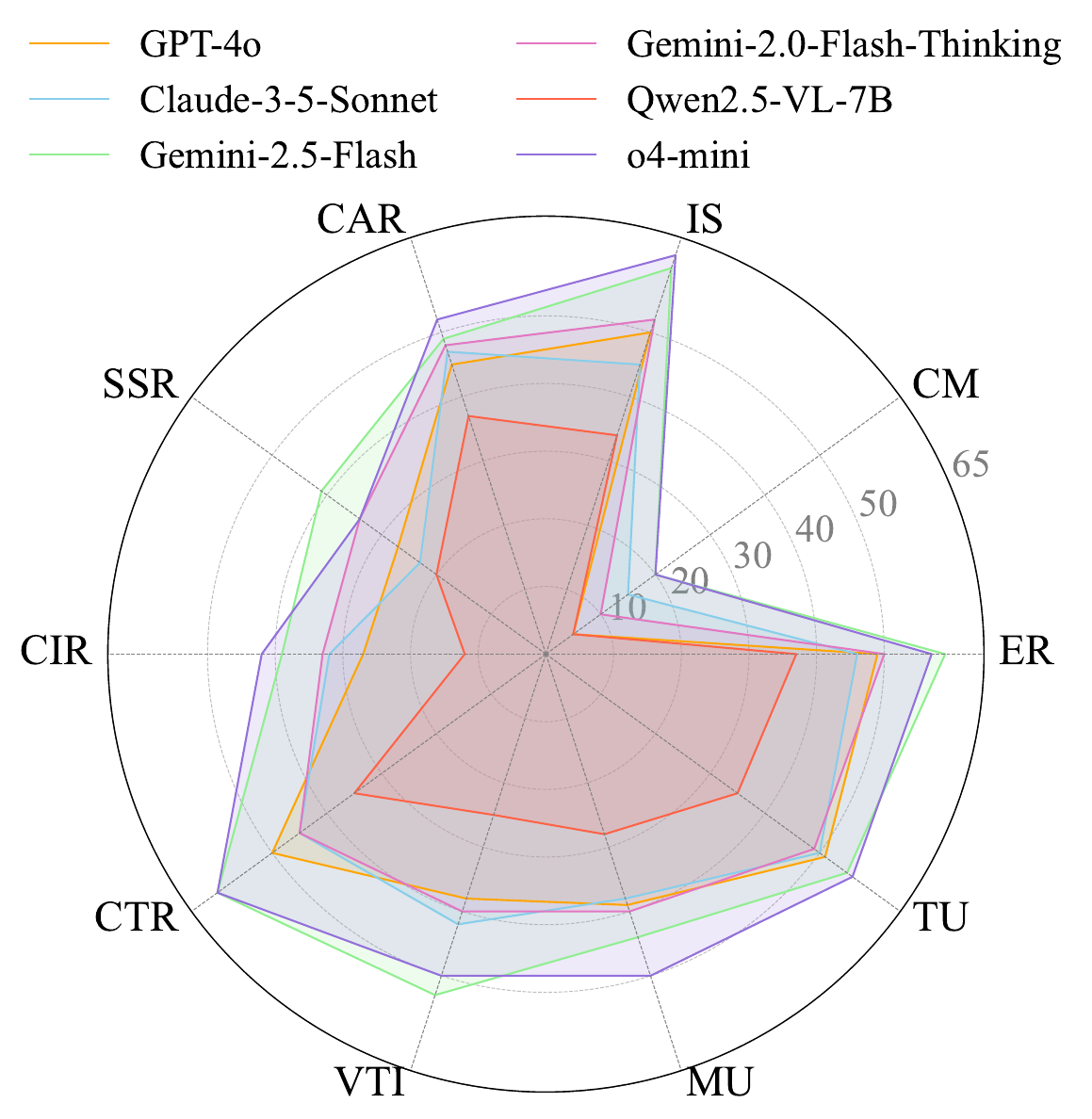}
  \caption{Performance on different tasks.}
  \label{fig:task analysis}
\vspace{-13pt}
\end{wrapfigure}
\textbf{Firstly, the models performed better on implicit tasks than on explicit tasks} (with an average gain of +7.9\%). Through analysis of tasks and model responses, we found that in implicit tasks, video creators often embed implicit meanings throughout the entire video, resulting in abundant visual cues that can support reasoning. This reduces the requirements for multi-modal reasoning and clue localization. In contrast, explicit tasks demand finer-grained reasoning and the ability to identify specific evidence. For example, in the implicit task at the bottom of Figure \ref{fig:benchmark_intro}, many frames provide clues suggesting that the girl symbolizes good luck (e.g., room number, flowers, lighting, weather, etc.). In contrast, the explicit task at the top contains only a few key frames where the hidden lighter in magician's hand can be seen.

\textbf{Secondly, the models performed particularly poorly on \textit{Counterintuitive Reasoning (CIR)}, \textit{Sequential Structure Reasoning (SSR)}, and \textit{Comment Matching (CM)} tasks}. For CIR and SSR, poor performance mainly stems from the limited ability of models to perform multi-frame reasoning. These two tasks require the model to reason on long-range videos, rather than relying on internal knowledge. However, instead of analyzing to locate evidences in other frames, models often rely on surface-level visual perception of the question frame, followed by textual reasoning over question and options.
For CM, the results highlight a significant gap between model and human capabilities in implicit reasoning. While humans can infer underlying information such as humor and emotion with minimal cognitive effort~\citep{krishna-etal-2022-rankgen}, current models consistently fail to capture such subtleties. 

\textbf{Human Performance.} Humans achieved an average score of 86\%, highlighting a significant human-model gap. Although studies suggest models reached human-level performance on text tasks~\citep{guo2025deepseek, openai2023gpt4}, they still lag on multimodal reasoning. Humans can identify clues in videos easily, while models tend to focus on question frames rather than exploring other evidence frames. Specially, unlike models, humans perform slightly worse on implicit tasks, which is mainly due to the challenges posed by highly abstract implicit understanding in art and philosophy.





\subsection{Influence of Frames Count}
\label{sec:Frames}

\begin{wrapfigure}{r}{0.43\textwidth}
  \centering
  \vspace{-25pt}
  \includegraphics[width=0.43\textwidth]{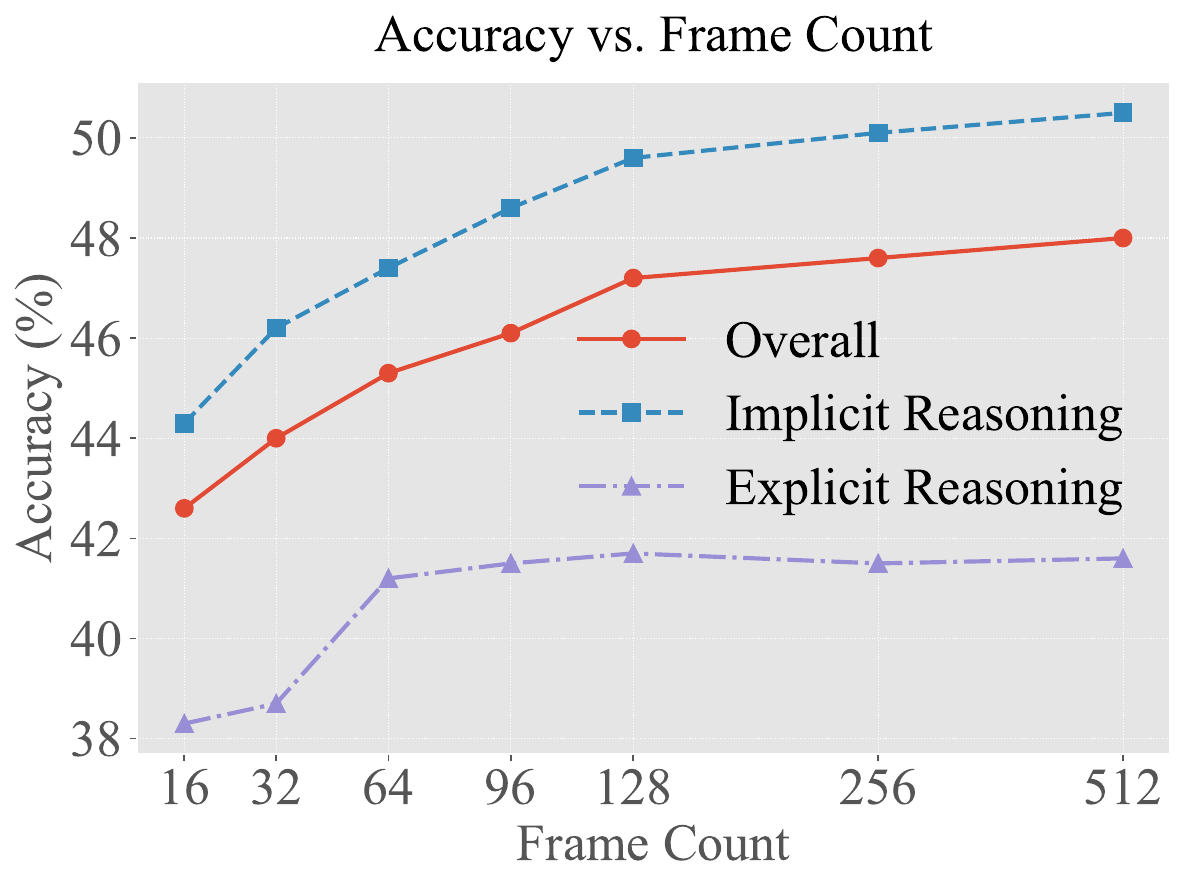}
  \caption{Accuracy with the increase of input frame counts.}
\vspace{-22pt}
\label{fig:Frames}
\end{wrapfigure}
For Gemini-2.0-Flash, which supports long video inputs, we evaluated performance changes as the number of frames increases. As shown in Figure \ref{fig:Frames}, accuracy improves with more frames, but the rate of improvement gradually slows. After sampling and observing the CoTs, it is found that the initial gains come from the addition of evidence frames, while the slowdown is mainly due to limited multi-frame reasoning ability of the model. Performance on implicit tasks continues to improve in later stages, as visual cues for such tasks are often dispersed throughout the video (as discussed in Section \ref{sec:diff-tasks}); more frames tend to provide more clues. In contrast, explicit clues are fewer and more localized.

\subsection{Influence of Audio Input}
\label{sec:Audio}


For models supporting full-modal input, we compared performance before and after incorporating audio. As shown in Table~\ref{table: audio}, overall performance improved with the addition of audio. Specifically, Gemini 2.0-Flash, Gemini 2.0-Flash-Thinking, and Phi-4-multimodal-instruct improved by 1.4\%, 1.0\%, and 1.0\%, respectively. This suggests that advancing research on fully multimodal models is a promising direction. A case study illustrating how audio aids reasoning is provided in Appendix~\ref{appendix:audio}.

\begin{table*}
\vspace{-5pt}
  \centering  
  \setlength\tabcolsep{3pt}
  \fontsize{7pt}{10pt}\selectfont
\caption{\label{table: audio}
    The impact of adding audio modality on the performance (accuracy \%) on different tasks.
  }
\begin{tabular}{lccccccccc}
\toprule
 &  & \multicolumn{2}{c}{\textbf{Tasks}} & \multicolumn{6}{c}{\textbf{Categories}} \\
 \cmidrule(lr){3-4} \cmidrule(lr){5-10}
 & Overall & \textit{Imp.} & \textit{Exp.} & \textit{Art} & \textit{Life} & \textit{TV} & \textit{Film} & \textit{Ani.} & \textit{Phi.} \\
\midrule
 Gemini-2.0 & 42.6 & 44.3 & 38.3 &30.9 & 32.2 & 40.7 & 40.6 & 58.5 & 24.4 \\
~+audio & $44.0^{ \textcolor{darkred}{\uparrow 1.4}}$ & $46.2^{ \textcolor{darkred}{\uparrow 1.9}}$ & $38.3^{ {- 0.0}}$ & $31.0^{ \textcolor{darkred}{\uparrow 0.1}}$ & $31.6^{ \textcolor{darkblue}{\downarrow 0.6}}$ & $42.3^{ \textcolor{darkred}{\uparrow 1.6}}$ & $41.0^{ \textcolor{darkred}{\uparrow 0.4}}$ & $61.1^{ \textcolor{darkred}{\uparrow 2.6}}$ & $29.1^{ \textcolor{darkred}{\uparrow 4.7}}$ \\
Gemini-2.0-thinking & 45.0 & 46.6 & 40.6 & 34.5 & 31.6 & 38.6 & 48.3 & 60.1 & 25.6 \\
~+audio & $46.0^{ \textcolor{darkred}{\uparrow 1.0}}$ & $48.4^{ \textcolor{darkred}{\uparrow 1.8}}$ & $39.7^{ \textcolor{darkblue}{\downarrow 0.9}}$ & $31.7^{ \textcolor{darkblue}{\downarrow 2.8}}$ & $33.9^{ \textcolor{darkred}{\uparrow 2.3}}$ & $44.4^{ \textcolor{darkred}{\uparrow 5.8}}$ & $42.7^{ \textcolor{darkblue}{\downarrow 5.6}}$ & $62.4^{ \textcolor{darkred}{\uparrow 2.3}}$ & $32.6^{ \textcolor{darkred}{\uparrow 7.0}}$ \\

Phi-4-multimodal-instruct & 26.7 & 29.4 &  19.4 & 19.4 & 19.2 & 25.9 & 26.4 & 33.9 & 24.4 \\
~+audio & $27.7^{ \textcolor{darkred}{\uparrow 1.0}}$ & $31.3^{ \textcolor{darkred}{\uparrow 1.9}}$ & $18.1^{ \textcolor{darkblue}{\downarrow 1.3}}$ & $15.4^{ \textcolor{darkblue}{\downarrow 3.0}}$ & $19.7^{ \textcolor{darkred}{\uparrow 0.5}}$ & $24.5^{ \textcolor{darkblue}{\downarrow 1.4}}$ & $27.8^{ \textcolor{darkred}{\uparrow 1.4}}$ & $37.3^{ \textcolor{darkred}{\uparrow 3.4}}$ & $26.7^{ \textcolor{darkred}{\uparrow 2.3}}$ \\

\bottomrule
\end{tabular}
\vspace{-10pt}
\end{table*}




\subsection{Error Analysis}
\label{sec:error analysis}

\begin{wrapfigure}{r}{0.4\textwidth}
\vspace{-43pt}
  \centering
  \includegraphics[width=0.4\textwidth]{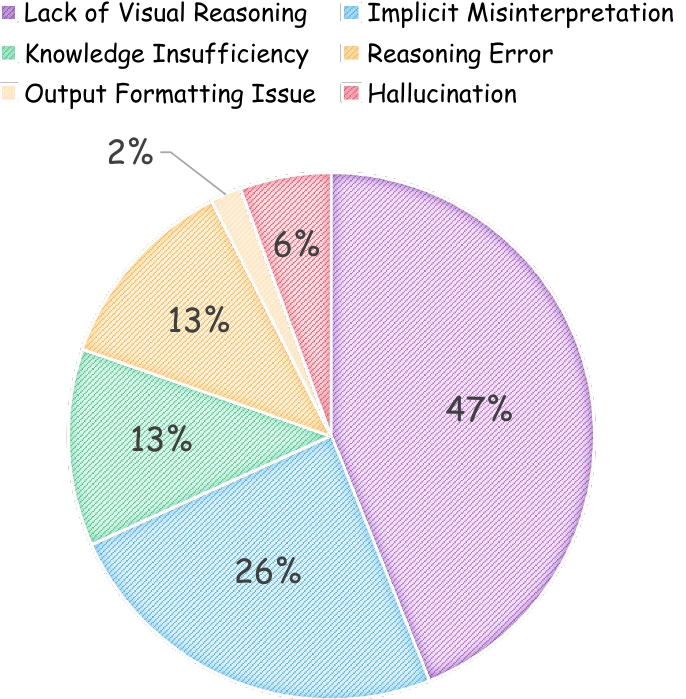}
  \caption{Error analysis of GPT-4o.}
  \label{fig:error analysis}
  \vspace{-15pt}
\end{wrapfigure}

We sampled 100 incorrect responses from GPT-4o for error analysis. 
The errors can be categorized as follows: \textbf{(1) Lack of Visual Reasoning:} the model often failed to locate the correct evidence frames and lack of long-range, multi-frame visual reasoning. \textbf{(2) Implicit Misinterpretation:} revealing a significant understanding gap between the model and human cognition. \textbf{(3) Knowledge Insufficiency:} the model lacks some intrinsic knowledge \textbf{(4) Reasoning Error:} during the multi-step deduction process. \textbf{(5) Hallucination:} the model introduced fake or unsupported information. \textbf{(6) Output Formatting Issue:} model refusals or formatting errors prevent answer extraction. Among error cases, \textit{Lack of Visual Reasoning} accounts for the largest proportion. This indicates that current models still lack genuine multimodal reasoning capabilities. They tend to rely on text-based reasoning after briefly perceiving frames adjacent to the question, rather than engaging in long-range, multi-frame video reasoning. Most existing reasoning models remain inadequate in integrating multimodal information into the reasoning process and performing thorough analysis. In contrast, models like GPT-5, o3 exhibits a better reasoning paradigm, as shown in Figure \ref{fig:good_vs_bad} for comparison. Details of the error category annotation pipeline are provided in Appendix \ref{appendix:error_analysis_qa}. 

\begin{figure}[t]
    \centering
    \includegraphics[width=\linewidth]{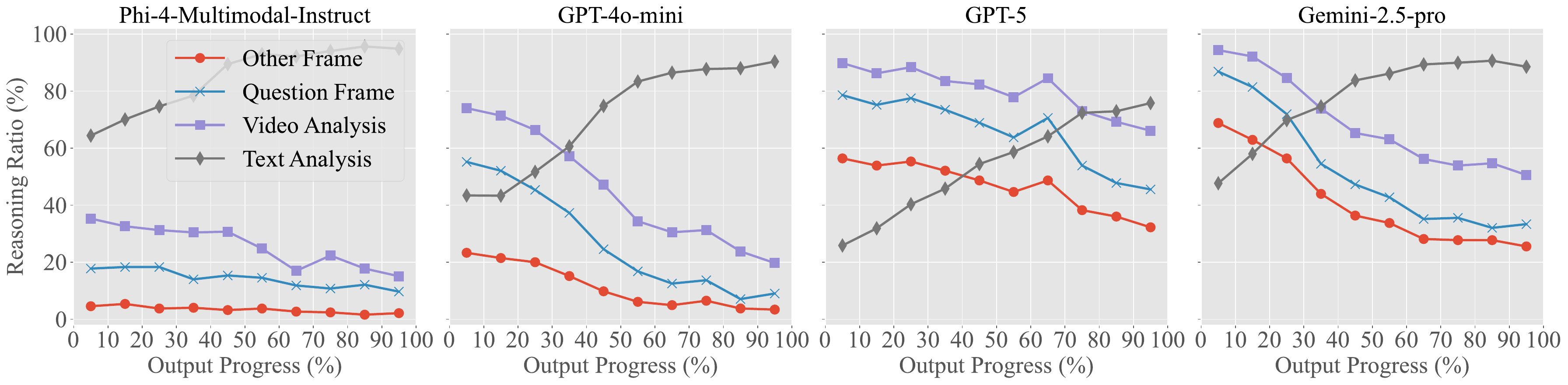}
    \caption{CoT content across different stages. The y-axis indicates the ratio of the 500 sampled CoTs that include analysis of these four types of content at each stage.}
    \label{fig:CoT_analysis}
\vspace{-1em}
\end{figure}

We further analyzed model CoTs to examine performance differences by categorizing each reasoning step as either video or text analysis (e.g., options), with video analysis further divided into question-frame and other-frame analysis (details in Appendix~\ref{appendix:cot_analysis}). We sampled 500 CoTs from models, split each into 10 equal-length segments, and used GPT-4.1 to label each segment. As shown in Figure \ref{fig:CoT_analysis}, where models further to the right perform better on MMR-V, models with better performance on MMR-V show more video analysis, especially on \textbf{other frames} (red line). Notably, GPT-5 and Gemini-2.5-pro stand out with strong analysis of non-question frames, highlighting the value of enhanced multi-frame visual reasoning in video reasoning tasks.


\section{Related Work}

\textbf{Video Understanding Benchmark.} Existing video benchmarks mainly evaluate models' perception and understanding of video, such as action recognition~\citep{khattak2024good, mangalam2023egoschema, patraucean2023perception, xiao2021next} and description~\citep{xu2017video, xu2016msr}. Recent works like Video-MME~\citep{fu2024video}, MVBench~\citep{li2024mvbench}, TempCompass~\citep{liu2024tempcompass} and MMBench-Video~\citep{fang2024mmbench} extend video understanding to diverse video types, enabling more comprehensive assessments. Benchmarks such as LVBench~\citep{wang2024lvbench}, LongVideoBench~\citep{wu2024longvideobench}, and CG-Bench~\citep{chen2024cg} further introduce long video QA. However, these largely test whether models can extract relevant information from long videos, with evaluation remaining perception-oriented. MMR-V instead assesses whether models can perform multi-frame, long-span, multimodal autonomous reasoning from questions.

\textbf{Multimodal Reasoning. }Recent advancements has greatly enhanced LLM reasoning~\citep{guo2025deepseek, jaech2024openai, team2025kimi, zhao2024marco}, yet the evaluations still focus on text-based reasoning~\citep{hendrycks2021measuring, bai2024longbench, gsm8k, rein2024gpqa, wang2024mmlu}. MLLMs still lack thorough assessment in this area~\citep{qin2023good, wang2023large}. Current multimodal reasoning benchmarks mainly involve mathematical or coding tasks in image form~\citep{wang2024measuring, shi2024math, ying2024mmt}, which primarily test visual recognition followed by text reasoning. True multimodal reasoning requires richer cross-modal cues. To address this gap, MMR-V is designed to evaluate video-based reasoning tasks.

\textbf{Video Reasoning.} Recently, several benchmarks have emerged to access reasoning in videos. VideoEspresso~\citep{han2025videoespresso} utilizes a fully automated pipeline to generate scalable CoT reasoning data in videos. Video-MMMU~\citep{hu2025video} and MMVU~\citep{zhao2025mmvu} focus on evaluating specialized knowledge reasoning within multi-disciplinary videos (e.g., physics, economics). VRBench~\citep{yu2025vrbench} targets narrative reasoning in long narrative videos and designs a human-model collaborative stepwise annotation pipeline. VideoReasonBench~\citep{liu2025videoreasonbench} assesses vision-centric reasoning using synthetic videos generated by engines. MMR-V distinguishes itself by ensuring all QA pairs are purely manually annotated, evaluating general multimodal reasoning across diverse real-world open-domain videos.

\section{Conclusion}
\label{conclusion}

This paper introduces \textbf{MMR-V: A Benchmark for Multimodal Deep Reasoning in Videos}. All tasks are human-annotated. MMR-V poses a significant challenge for current models, with the best performance still trailing human accuracy by 21.7\%. This highlights a human-model gap in interpreting and reasoning about video information. Notably, we observe that models with higher accuracy on MMR-V tend to perform more extensive and in-depth analysis of videos, suggesting that incorporating multi-frame reasoning into CoT or leveraging tool use may be promising directions for advancing video reasoning. We hope MMR-V will serve as a reliable evaluation benchmark for the development of MLLMs and offer valuable insights into advancing multimodal reasoning research.


\section*{Acknowledgements} This work is supported by the National Natural Science Foundation of China (No. U24A20335, No. 62406321). This work is also supported by Beijing Natural Science Foundation (L243006).

\section*{Ethics Statement}
\label{Ethics statement}
All experimental procedures involving human participants were conducted in accordance with the relevant ethical guidelines. The MMR-V data was gathered and annotated by compensated internal annotators, from whom all necessary informed consent was obtained. The dataset was carefully reviewed for safety and potential biases. We have prepared the dataset for public release under the CC-BY 4.0 license. 

\section*{Reproducibility Statement}
\label{Reproducibility statement}

We have taken several steps to enhance the reproducibility of our research. The code for the experiments and analyses reported in the main text is provided in the supplementary materials. The experimental setup of the main experiments and the selection of video frames are detailed in Section~\ref{sec:exp_setup} and Appendix~\ref{sec: exp details}. Further details regarding dataset construction and preprocessing are described in Appendix~\ref{sec:construction} and Appendix~\ref{sec:task details}. Finally, our dataset has been prepared for open release to support future research. The JSON file containing the questions, answers, and related information for MMR-V can be found in the supplementary materials.





\bibliography{iclr2026_conference}
\bibliographystyle{iclr2026_conference}

\appendix
\section{LLM USAGE STATEMENT}

In this paper, LLMs were used to check grammar and polish text, and we carefully verified that their use did not alter the original meaning. In addition, AI assistants were employed to improve the presentation of some tables and figures, but we ensured the experimental results remained unchanged.

\section{Limitations}
\label{sec:limitations}

Despite our efforts to improve our work, several limitations remain. (1) Scaling MMR-V is challenging due to the high cost of manual annotation and verification, as all tasks and correct answers are curated and reviewed by human annotators. (2) Although we strive to cover a wide range of video and task types, certain real-world categories (such as mystery, puzzle-solving, and gaming) are still underrepresented. (3) The majority of videos in MMR-V are in English, with only a small proportion in other languages such as Chinese, French, Thai, and German, which constrains its multilingual applicability. We will further study and try to solve this issue in the future.

\section{MMR-V Construction}
\label{sec:construction}
\subsection{Checklist}
According to the MMR-V construction principles introduced in the main text Section \ref{principles} , we wrote the following annotation checklist:
\begin{tcolorbox}[size=title,opacityfill=0.05,breakable]
(1) You are expected to \textbf{watch the entire video} before formulating any questions or answers. 

(2) Each question must require \textbf{long-distance, multi-frame reasoning} and cannot be answered through direct perception (ensuring compliance with Principles 1 and 2). 

(3) To ensure \textbf{consistency with real-world user} perception (Principle 3), annotators are encouraged to refer to the official interpretation of the original video author and user consensus (highly praised comments in the comment section) when writing or verifying the correct answer. This helps mitigate annotator bias and ensures that the reasoning task reflects the understanding of a wider audience.
\end{tcolorbox}
\begin{figure}[t]
    \centering
    \includegraphics[width=\linewidth]{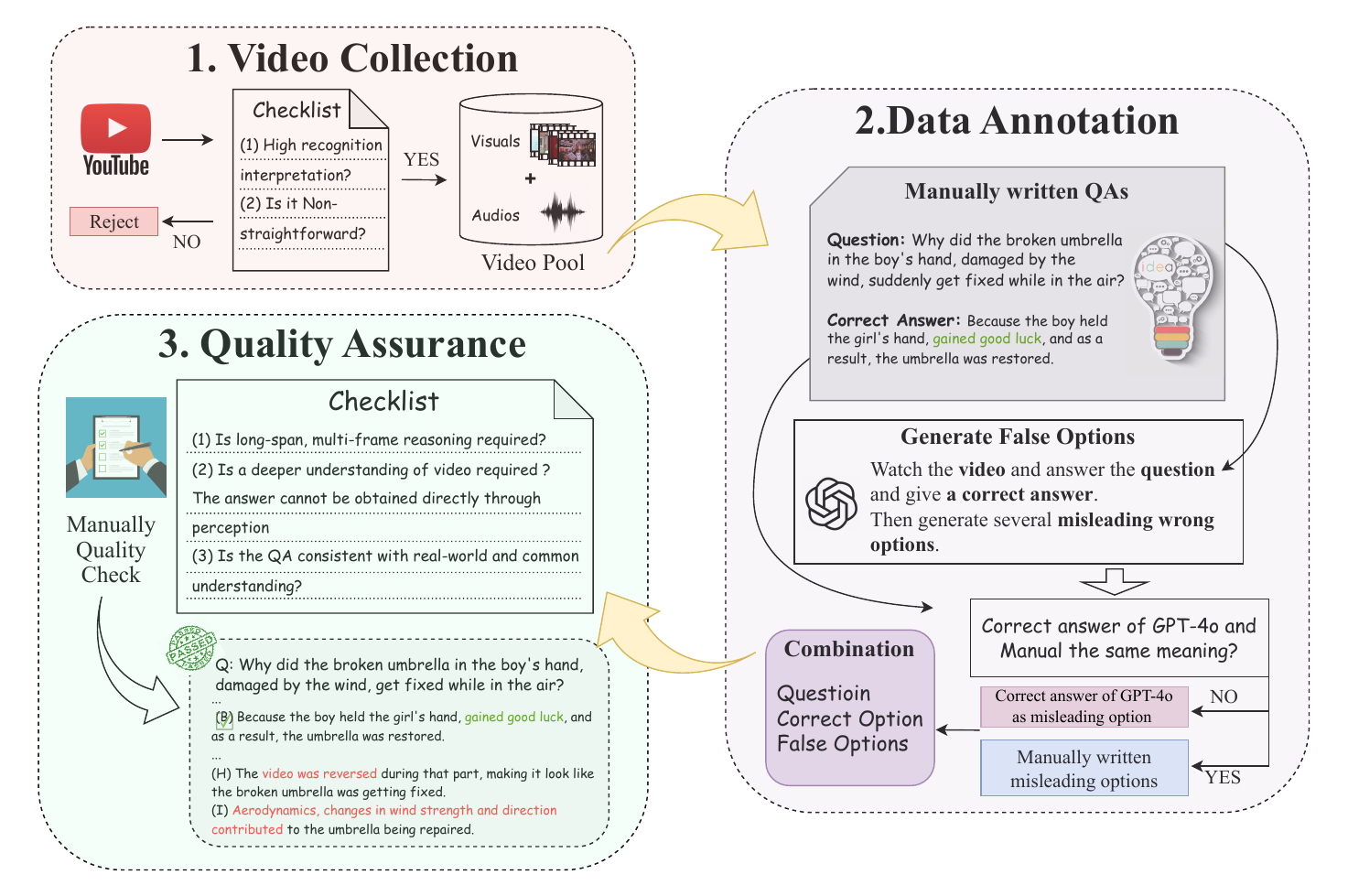}
    \caption{MMR-V Construction Pipeline.}
    \label{fig:pipeline}
\vspace{-1.6em}
\end{figure}

\begin{figure}[t]
    \centering
    \includegraphics[width=\linewidth]{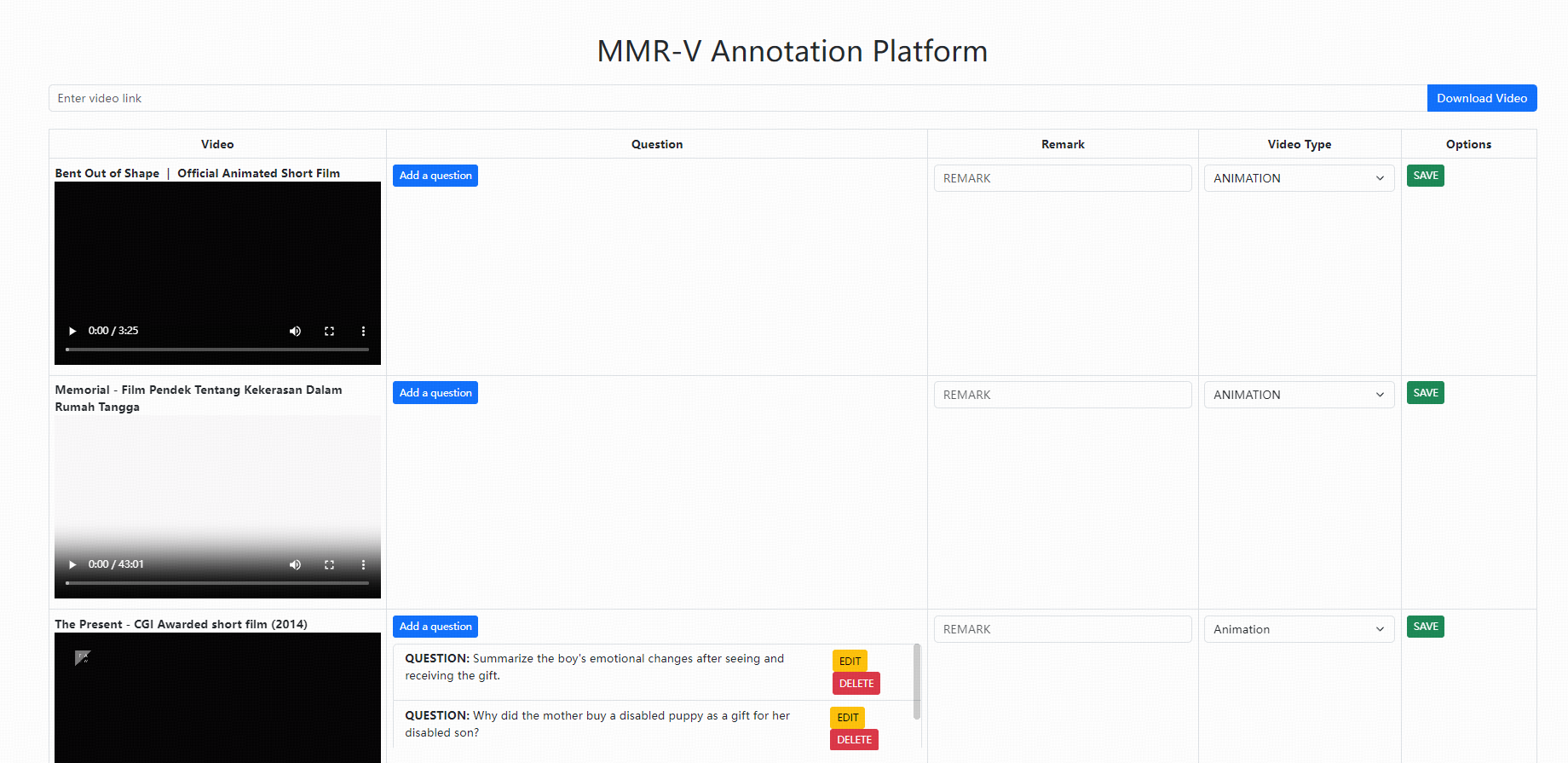}
    \caption{Annotation Platform of MMR-V.}
    \label{fig:platform}
\vspace{-1em}
\end{figure}

\subsection{Construction Pipeline}
In this section, we present the construction process of MMR-V Bench in a macro sense. The whole process is divided into three stages: \textbf{video collection}, \textbf{data annotation}, and \textbf{quality assurance}. For \textbf{video collection}, we designed a checklist to ensure the quality and diversity of videos in the Bench. "High recognition interpretation?" ensures that the questions raised and the annotated answers based on the video have references that are consistent with public cognition (official interpretations or highly praised comments) to alleviate the subjective bias of the annotator. "Is it Non-straightforward?" ensures that the video is not a straightforward narrative, which is conducive to increasing the reasoning difficulty of the question. For \textbf{data annotation}, as described in section \ref{sec:Data Annotation} of the main text, we use gpt-4o to assist in annotation with interference options. Let the model generate the correct answer based on the question, and manually review to ensure that the correct answer generated by the model is different from the manual annotation. If they are different, the answer generated by gpt-4o is used as the interference item, otherwise the interference item is manually written. For \textbf{quality assurance}, we designed a checklist for human reviewers to check the correctness and difficulty of the tasks. The annotation platform is shown in Figure \ref{fig:platform}.

\subsection{Pipeline of Error Analysis}
\label{appendix:error_analysis_qa}

To ensure the reliability of the error analysis presented in Section \ref{sec:error analysis}, we implemented an annotation pipeline and quality control strategy. Firstly, we categorized model failures into six distinct types based on initial analysis of the false answers. To standardize the criteria, we selected representative anchor examples for each category to distinguish boundary cases. Some examples are presented in Appendix \ref{sec:case}. Next, to mitigate author bias, we invited non-author evaluators to review the definitions and anchor examples, ensuring they align with general human understanding before formal annotation. Finally, the primary authors annotated the sampled responses based on the established guidelines. Any disagreements were discussed and resolved to ensure 100\% consensus of authors on the final error distribution.

 \begin{figure}
  \centering
  \begin{subfigure}[b]{0.48\textwidth}
      \centering
      \includegraphics[width=\textwidth]{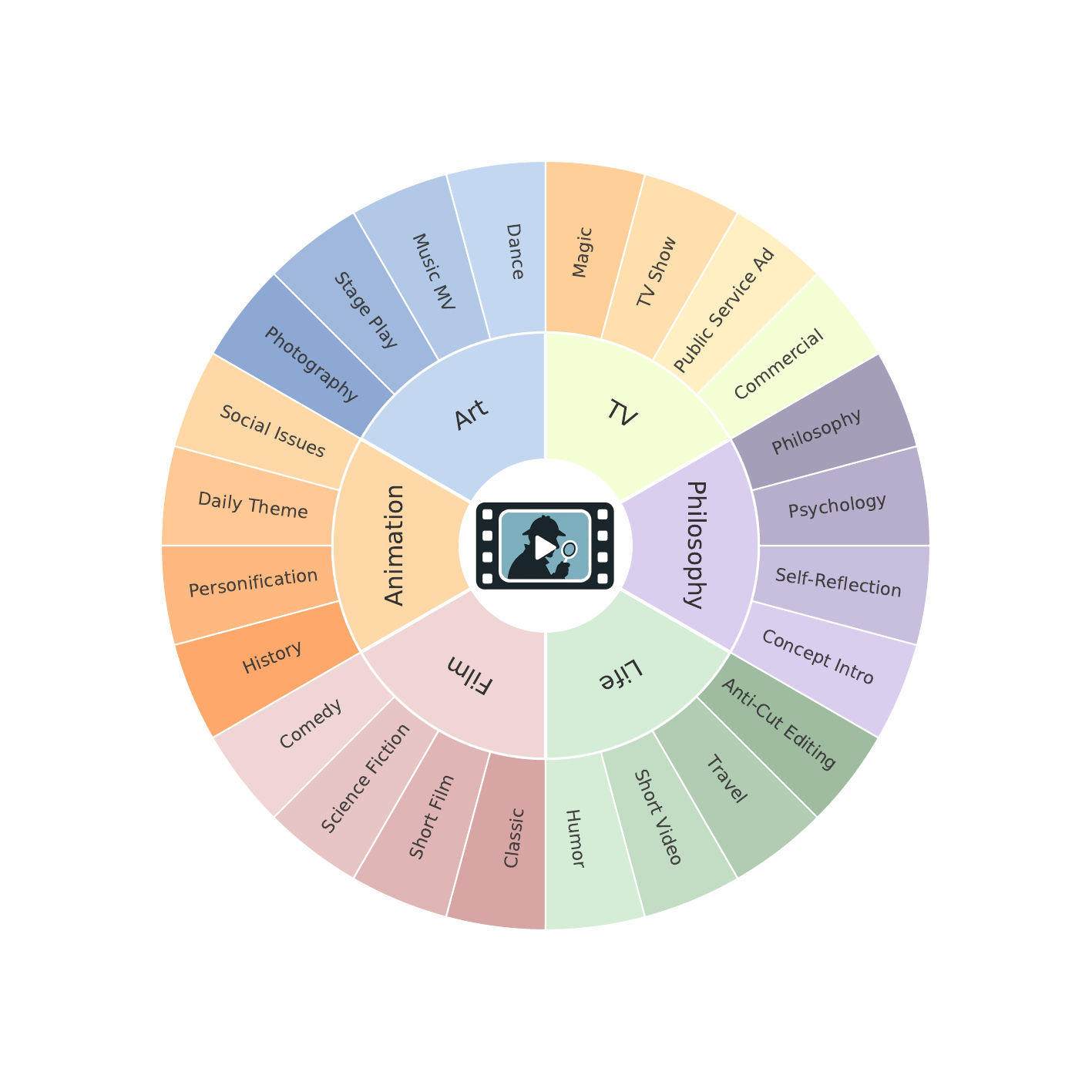}
      \caption{Video categories.}
      \label{fig:video-categories}
  \end{subfigure}
  \begin{subfigure}[b]{0.48\textwidth}
      \centering
      \includegraphics[width=\textwidth]{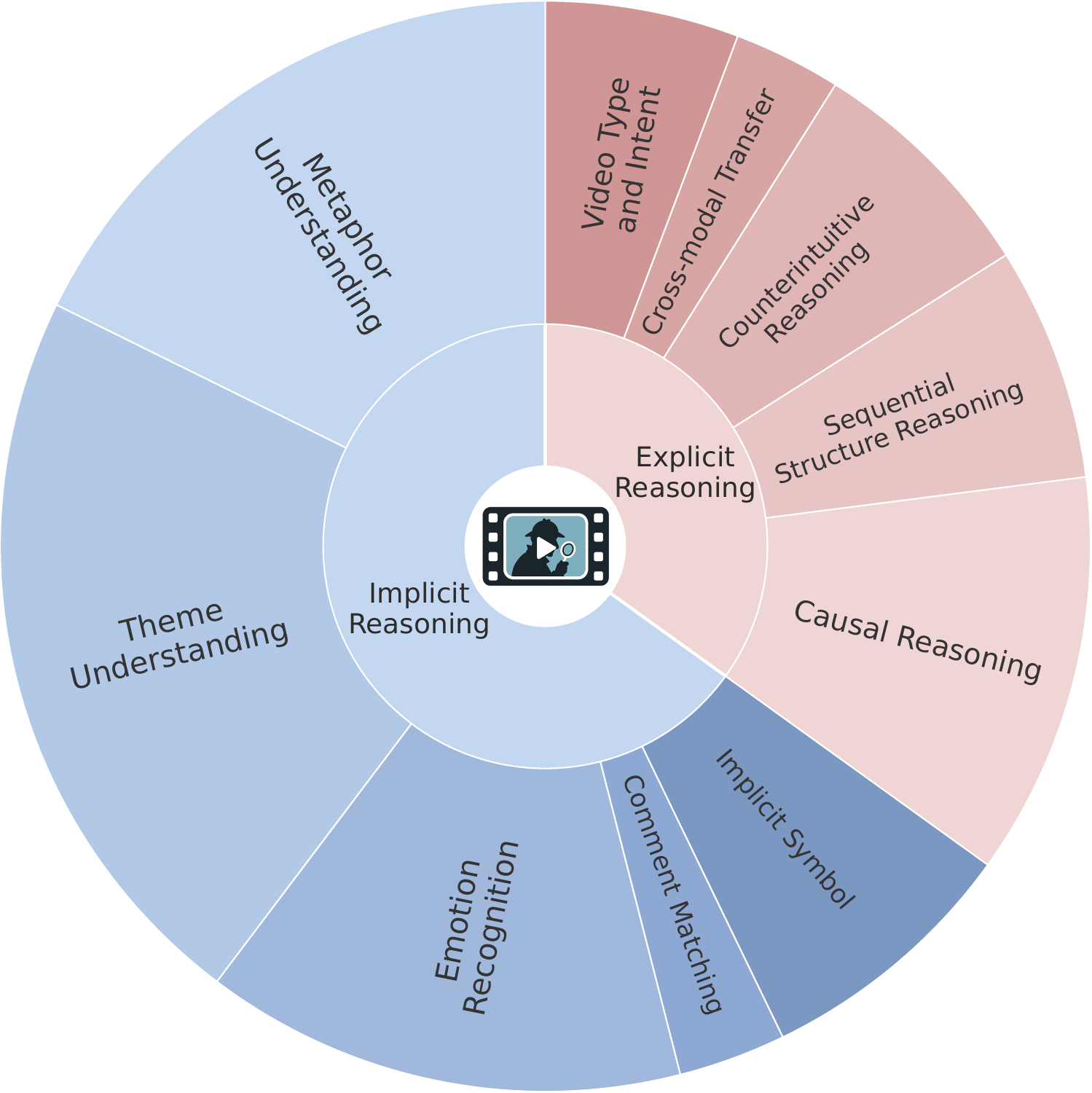}
      \caption{Proportion of different tasks.}
      \label{fig:ability-type}
  \end{subfigure}

\caption{(a) Video categories in MMR-V Bench. (b) Proportion of different tasks in MMR-V Bench.}
\label{figure:data-statistic}
\end{figure}

\section{Diversity of MMR-V}
\label{sec:diversity}
In this section, we show the diversity of MMR-V Bench, including video diversity and task diversity. For video, we show the six categories of videos in MMR-V in Figure \ref{fig:video-categories}, including  Life, Animation, Film, Art, TV, and Philosophy. At the same time, for each category, we divide it into several subcategories to better understand the classification of video categories. Secondly, in section \ref{sec:statistics} we show the diversity of video length, ranging from 7 seconds to 3771 seconds. For tasks, we divide them into two parts, ten categories and 33 subcategories, three levels. The division of the first and second levels, as well as the proportion of different types of tasks, can be seen in Figure \ref{fig:ability-type}.

\begin{table*}[t]
\centering
\small
\renewcommand{\arraystretch}{1.2}
\setlength{\tabcolsep}{4pt}
\caption{Comparison with existing video benchmarks. \textbf{\#Dur.}: Average video duration in seconds. \textbf{QA Anno.}: Annotation method (A: Automatic, M: Manual, A\&M: Both).}
\resizebox{0.95\linewidth}{!}{%
\begin{tabular}{l|c|c|c|l|c|c|c}
\hline
\textbf{Benchmark} & \textbf{\#Videos} & \textbf{\#Dur.(s)} & \textbf{\#QA Pairs} & \textbf{Data Source} & \textbf{Task Categories } & \textbf{Multilingual} & \textbf{QA Anno.} \\
\hline
MVBench & 3,641 & 16.0 & 4,000 & Open-Domain & 20 & $\times$ & A \\
EgoSchema & 5,063 & 180.0 & 5,063 & Egocentric & $\times$ & $\times$ & A \\
TempCompass & 410 & 11.4 & 7,540 & Open-Domain & 11 & $\times$ & A\&M \\
Video-MME & 900 & 1017.9 & 2,700 & Open-Domain & 12 & \checkmark & M \\
LongVideoBench & 3,763 & 473.0 & 6,678 & Open-Domain & 17 & $\times$ & M \\
LVBench & 103 & 4101.0 & 1,549 & Open-Domain & 26 & $\times$ & M \\
CGBench & 1,219 & 1624.4 & 12,129 & Open-Domain & 12 & $\times$ & M \\
Video-MMMU & 300 & 506.2 & 900 & Open-Domain & 6 & $\times$ & M \\
MMVU & 1,529 & 51.4 & 3,000 & Multi-Disc & $\times$ & $\times$ & M \\
VRBench & 960 & 5796.0 & 8,243 & Narrative & 7 & \checkmark & A\&M \\
\hline
\rowcolor{gray!15} \textbf{MMR-V (Ours)} & 317 & 277.0 & 1,257 & Open-Domain & 33 & \checkmark & M \\
\hline
\end{tabular}%
}
\label{tab:benchmark_comparison}
\end{table*}

\section{Task Details}
\label{sec:task details}
The tasks in MMR-V can be divided into three levels. Level 1: Implicit Reasoning \& Explicit Reasoning. Level 2: Contains ten task classes. Level 3: Contains 33 task subclasses. Next, we will introduce these tasks with some task examples\footnote{The videos linked here are the original source versions. Note that the actual videos in MMR-V were processed to trim frames containing shortcuts to the answers.}.


\begin{table*}
  \centering  
  \setlength\tabcolsep{5pt}
  \fontsize{8pt}{11pt}\selectfont
\caption{\label{table: task_level}
    Three-level classification of tasks in MMR-V.
  }

\begin{tabular}{c|l|l}
\toprule
Ability Type L1                     & \multicolumn{1}{c|}{Ability Type L2} & \multicolumn{1}{c}{Ability Type L3}                                                                                                  \\ \midrule
\multirow{5}{*}{Implicit Reasoning} & Metaphor Understanding (MU)          & \begin{tabular}[c]{@{}l@{}}Structural Metaphor, Orientational Metaphor,\\ Ontological Metaphor, Creative Metaphor\end{tabular}        \\ \cline{2-3} 
                                    & Theme Understanding (TU)             & \begin{tabular}[c]{@{}l@{}}Philosophical Concepts, Social Issues, Personal Reflection,\\ Everyday Topics, Video Naming\end{tabular}   \\ \cline{2-3} 
                                    & Emotion Recognition (ER)             & \begin{tabular}[c]{@{}l@{}}Explicit Emotion, Implicit Emotion,\\ Meta-emotion, Audience Emotion\end{tabular}                          \\ \cline{2-3} 
                                    & Comment Matching (CM)                & Humorous, Thought-provoking, Trending                                                                                                   \\ \cline{2-3} 
                                    & Implicit Symbol (IS)                 & Cultural Symbols, Art Symbols, Other Symbols                                                                                            \\ \midrule
\multirow{5}{*}{Explicit Reasoning} & Causal Reasoning (CAR)               & Forward Reasoning, Backward Reasoning                                                                                                  \\ \cline{2-3} 
                                    & Sequential Structure Reasoning (SSR) & \begin{tabular}[c]{@{}l@{}}Narrative Structure, Core Connecting Elements,\\ Inference on Editing Techniques, Hallucination\end{tabular} \\ \cline{2-3} 
                                    & Counterintuitive Reasoning (CIR)     & \begin{tabular}[c]{@{}l@{}}Magic Deconstruction or Special Effects Editing,\\ Artistic Techniques, Humor and Exaggeration\end{tabular} \\ \cline{2-3} 
                                    & Cross-modal Transfer Reasoning (CTR) & Video-to-Text, Video-to-Audio, Video-to-Video                                                                                          \\ \cline{2-3} 
                                    & Video Type and Intent (VTI)          & Video Type, Video Intent                                                                                                               \\ \bottomrule
\end{tabular}
\end{table*}

\subsection{Implicit Reasoning Tasks}

\subsection*{\textcolor{darkblue}{I. Metaphor Understanding (MU)}}

For the definition of subclasses of the metaphor understanding task, we mainly refer to the book Metaphors We Live~\citep{lakoff2008metaphors} By by George Lakoff and Mark Johnson, which introduces metaphor-related concepts in detail.

\subsection*{\textcolor{darkblue}{I.1. Structural Metaphor}}

\textbf{\textcolor{darkred}{Task Description}}: There are structural similarities between the subject and object. For example, time can be compared to flowing water, both of which have the structure of flow and passing away.\\
\textbf{\textcolor{darkred}{Example Question}}: 

\begin{tcolorbox}[size=title,opacityfill=0.05,breakable]
\textbf{Question}: What does the brown coat in the video symbolize?

\textbf{Options}: 

 (A) It is said to represent the family's long - lost fortune that they are still searching for.

(B) The brown coat symbolizes the lost hope of the family because it was worn during a difficult time.

(C) It refers to a coat that has been washed and taken out to dry, likely worn by the father.",

(D) It symbolizes the father in a family, who protects his family in times of difficulty.

(E) It represents the fear of the outside world.

(F) The unfulfilled dreams of the children in the family as they always saw it as a sign of something unattainable.

(G) The brown coat in the video represents a raincoat, used to protect the clothes inside from getting wet.

(H) The bad luck that has been following the family for generations.

\textbf{CorrectAnswer}: (D)

\textbf{Video}: father - 1 minute emotional award winning - \href{https://www.youtube.com/watch?v=fUR5ks-t9TI}{video\_url} 
\end{tcolorbox}

\subsection*{\textcolor{darkblue}{I.2. Orientational Metaphor}}
\label{Orientational Metaphor}
\textbf{\textcolor{darkred}{Task Description}}: There are similarities in direction or composition between the subject and the metaphor, for example, walking up a staircase is compared to ambition.\\
\textbf{\textcolor{darkred}{Example Question}}: 

\begin{tcolorbox}[size=title,opacityfill=0.05,breakable]
\textbf{Question}: Why does the dance, which is filled with artistry and beauty throughout, end with a descent?

\textbf{Options}:

(A) There is a connection between the fall and the creation at some point.

(B) It represents a dive to explore new depths, both literal and metaphorical.

(C) It indicates the dancer's exhaustion, capturing a moment of fatigue.

(D) It reflects the calmness of the ocean, evoking a sense of tranquility.

(E) It highlights the theme of rebirth, symbolizing renewal and transformation.

(F) It represents the beauty of underwater life, showcasing its unique allure.
            
(G) It symbolizes being weighed down by emotions, expressing inner turmoil.

(H) It symbolizes the end of a dream, marking a moment of conclusion.

(I) It shows the dancer's connection to water, emphasizing fluidity and grace.

(J) It symbolizes a return to nature and surrender to life's forces, embracing the natural flow.

(K) It signifies the end of the dance's energy, indicating a point of culmination.

\textbf{CorrectAnswer}: (A)

\textbf{Video}: Falling - Underwater dance
 - \href{https://www.youtube.com/watch?v=GucoPpK8g6g}{video\_url} 
\end{tcolorbox}

\subsection*{\textcolor{darkblue}{I.3. Ontological Metaphor}}
\textbf{\textcolor{darkred}{Task Description}}: This metaphor involves viewing an abstract concept as a concrete entity. Usually, the core concept of the entire video is turned into a concrete entity to tell the story.\\
\textbf{\textcolor{darkred}{Example Question}}: 

\begin{tcolorbox}[size=title,opacityfill=0.05,breakable]
\textbf{Question}: The scene around 1:00 metaphorically represents what aspect of communities?

        \textbf{Options}: 

            (A) Communities can build their resilience to setbacks by working together and adapting to new challenges.
            
            (B) Promoting individual success in competitive environments.
            
            (C) Building resilience through community partnerships.
            
            (D) Overcoming challenges for community progress.
            
            (E) Celebrating the individual achievements of community members.
            
            (F) Developing sustainable practices for environmental harmony.
            
            (G) Decision-making processes of a community.
            
            (H) The interconnectedness of global communities.
            
            (I) Isolation of communities for self-sufficiency.
            
            (J) The role of external aid in community development.
            
            (K) Highlighting the diversity of cultures within a community.
            
        \textbf{CorrectAnswer}: (A)

\textbf{Video}: Resilience: Anticipate, organise, adapt
 - \href{https://www.youtube.com/watch?v=yyX6UULJEic&t=60s}{video\_url} 
\end{tcolorbox}

\subsection*{\textcolor{darkblue}{I.4. Creative Metaphor}}
\textbf{\textcolor{darkred}{Task Description}}: This metaphor is usually carefully designed by the author for a specific video and needs to be understood in the context of the video.\\
\textbf{\textcolor{darkred}{Example Question}}: 

\begin{tcolorbox}[size=title,opacityfill=0.05,breakable]
\textbf{Question}: What is the pink fairy ball in the film?
        
        \textbf{Options}: 

            (A) It's a toy the boy picked up on the street, having no special connection to his condition.
            
            (B) They are the microorganisms in this world, living in every corner.
            
            (C) The pink fairy ball represents the boy's childhood dream of becoming a fairy.
            
            (D) It's a hallucination caused by lack of sleep, not related to antidepressants at all.
            
            (E) They are the boy's toys, which he bought to help treat his depression.
            
            (F) It is the effect of the antidepressants the boy is taking, which helps him see many things with vitality and positive effects.
            
            (G) It's an advertisement prop for a new product in the background of the scene.
            
            (H) The ball is a sign of the boy's wish to escape from his daily work routine.
            
            (I) The pink fairy ball is a symbol of the city's upcoming festival decorations

\textbf{CorrectAnswer}: (F)

\textbf{Video}: Soft Rain | Animated Short Film (2023)
 - \href{https://www.youtube.com/watch?v=tbsfc6Wn2DU&t=1s}{video\_url} 
\end{tcolorbox}

\subsection*{\textcolor{darkblue}{II. Theme Understanding (TU)}}
\subsection*{\textcolor{darkblue}{II.1. Philosophical Concepts}}
\textbf{\textcolor{darkred}{Task Description}}: The themes of the videos are usually about concepts and principles related to philosophy and psychology.\\
\textbf{\textcolor{darkred}{Example Question}}: 

\begin{tcolorbox}[size=title,opacityfill=0.05,breakable]
\textbf{Question}: What is the overall message that the animation aims to convey?
        
        \textbf{Options}:
            
            (A) It suggests happiness comes solely from financial achievements.
            
            (B) The animation emphasizes the need to avoid all responsibilities.
            
            (C) The animation aims to illustrate the ways to relieve stress.
            
            (D) It illustrates the mechanical process of water flow.
            
            (E) The animation encourages saving water to prevent wastage.
            
            (F) The animation conveys the importance of managing stress through self-care practices.
            
            (G) The animation highlights achieving success through hard work.

            (H) The animation suggests that ignoring stress leads to happiness.
        
            (I) The video underlines the significance of collective teamwork.
            
            (J) It depicts progress and growth through constant work.
        
        \textbf{CorrectAnswer}: (C)

\textbf{Video}: The Stress Bucket
 - \href{https://www.youtube.com/watch?v=uBl4g2SqZGY}{video\_url} 
\end{tcolorbox}

\subsection*{\textcolor{darkblue}{II.2. Social Issues}}
\textbf{\textcolor{darkred}{Task Description}}: The theme of the video is usually to reflect some problems existing in today's society and express a strong appeal of the author.\\
\textbf{\textcolor{darkred}{Example Question}}: 

\begin{tcolorbox}[size=title,opacityfill=0.05,breakable]

\textbf{Question}: What social reality does this video satirize?

\textbf{Options}: 

            (A) The rise of environmental awareness in urban settings.

            (B) The video represents the bystander effect in society.
            
            (C) The economic disparities in urban vs. rural areas.
            
            (D) The challenges of modern relationship dynamics.
            
            (E) The impact of fashion trends on daily life.
            
            (F) The increasing complexity of urban development planning.
            
            (G) The need for infrastructure improvement and road safety.
            
            (H) The influence of social media on public behavior.
            
            (I) The rapid pace of technological advancement in transportation.
            
            (J) The shift in societal values towards individualism.

\textbf{CorrectAnswer}: (B)

\textbf{Video}: Stone | 1 Minute Short Film | Hot Shot
 - \href{https://www.youtube.com/watch?v=5hPtU8Jbpg0}{video\_url} 
\end{tcolorbox}

\subsection*{\textcolor{darkblue}{II.3. Personal Reflection}}
\textbf{\textcolor{darkred}{Task Description}}: The author hopes that the video will inspire people to reflect on and resonate with things in their lives.\\
\textbf{\textcolor{darkred}{Example Question}}: 

\begin{tcolorbox}[size=title,opacityfill=0.05,breakable]

\textbf{Question}: What is the core concept that the film aims to convey?
        
        \textbf{Options}: 
        
            (A) Romantic relationships in adolescence.

            (B) The importance of education institutions.
            
            (C) Overcoming supernatural challenges.
            
            (D) The dynamics of family disagreements.
            
            (E) Exploration of technological advancement.
            
            (F) Not to judge others too quickly.
            
            (G) Journey of a superhero in saving the city.
            
            (H) Inter-species relations on Earth.
            
            (I) Power struggles in political leadership.
            
            (J) Historical recount of a famous personality.
        
        \textbf{CorrectAnswer}: (F)

\textbf{Video}: Award Winning SHORT FILMS Don't Judge | BATTI Hindi Heart Touching Short Movies | Content Ka Keeda
 - \href{https://www.youtube.com/watch?v=3-PXNnaatx4}{video\_url} 
\end{tcolorbox}

\subsection*{\textcolor{darkblue}{II.4. Everyday Topics}}
\textbf{\textcolor{darkred}{Task Description}}: The themes expressed in the videos are usually the sublimation of the insights and themes in daily life, such as praising maternal love, friendship, etc.\\
\textbf{\textcolor{darkred}{Example Question}}: 

\begin{tcolorbox}[size=title,opacityfill=0.05,breakable]

\textbf{Question}: What is implied by the contrast between the scenes around 0:47 and 1:11?

\textbf{Options}: 

            (A) The contrast shows that the mother is indecisive and can't make up her mind in a crisis.
            
            (B) It demonstrates the father's sense of responsibility and bravery, praising paternal love.
            
            (C) The contrast between the beginning and the end conveys a sense of tragedy, criticizing the destruction of the ecological environment by humans.
            
            (D) It shows that the father wants to abandon the child when facing danger.
            
            (E) It shows the bravery of the bird in the background, facing authority head-on, and praises courage.
            
            (F) The mother still protects her child at all costs even in the face of danger, which praises maternal love.
            
            (G) It implies that the father is doing it for self - preservation rather than out of love for the child.
            
            (H) It shows that even when there are many birds, they do not appear very united, and in the face of danger, they become a disorganized mess.
    
        \textbf{CorrectAnswer}: (F)

\textbf{Video}: Mother 1 minute Sad Emotional Award Winning Iranian Short Film Animation Animated
 - \href{https://www.youtube.com/watch?v=yXVs0So0mzc}{video\_url} 
\end{tcolorbox}

\subsection*{\textcolor{darkblue}{II.5. Video Naming}}
\textbf{\textcolor{darkred}{Task Description}}: Come up with a suitable title for this video or the core content of the video (dance, etc.). This tests the model's control over the overall content and whether it can get the subtleties of the title like humans.\\
\textbf{\textcolor{darkred}{Example Question}}: 

\begin{tcolorbox}[size=title,opacityfill=0.05,breakable]

\textbf{Question}: "Please come up with a suitable name for this dance.",

\textbf{Options}:
            
            (A) The Dance of the Butterfly.",
            
            (B) The Rhythm of the Phoenix.",
            
            (C) The Grace of the Swan.",
            
            (D) The Spirit of the Dragon.",
            
            (E) The Charm of the Peony.",
            
            (F) The Step of the Tiger.",
            
            (G) The Soul of Peacock",
            
            (H) The Beat of the Forest.",
            
            (I) The Leap of the Deer.",
            
            (J) The Spin of the Star.",
            
            (K) The Waltz of the Moon."

\textbf{CorrectAnswer}: (G)

\textbf{Video}: Yang Liping - The Soul of Peacock - Peacock Dance - Traditional Dance
 - \href{https://www.youtube.com/watch?v=241N7NGQgdM}{video\_url} 
\end{tcolorbox}

\subsection*{\textcolor{darkblue}{III. Emotion Recognition (ER)}}
\subsection*{\textcolor{darkblue}{III.1. Explicit Emotion}}
\textbf{\textcolor{darkred}{Task Description}}: Analyze the emotions of the characters in the video. Explicit emotions can usually be directly understood through facial expressions, body movements, etc.\\
\textbf{\textcolor{darkred}{Example Question}}: 

\begin{tcolorbox}[size=title,opacityfill=0.05,breakable]

\textbf{Question}: Summarize the boy's emotional changes between 6:00 and 7:00.
        
        \textbf{options}: 
            
            (A) Anger - Fear - Surprise and happiness
            
            (B) Sadness - Excitement - Helplessness
            
            (C) Disappointment - Let - down - Sorrow
            
            (D) Loneliness - Isolation - Solitude
            
            (E) Sadness - Grief - Mourning
            
            (F) Sadness - Shock - Surprise and happiness
            
            (G) Disappointment - Astonishment - Stupefaction
            
            (H) Loneliness - Isolation - Sorrow
            
            (I) Disappointment - Excitement - Helplessness
        
        \textbf{correctAnswer}: (F)

\textbf{Video}: CGI Animated Short Film: "Crunch" by Gof Animation | CGMeetup
 - \href{https://www.youtube.com/watch?v=iFUFFUb5W4w&t=16s}{video\_url} 
\end{tcolorbox}

\subsection*{\textcolor{darkblue}{III.2. Implicit Emotion}}
\textbf{\textcolor{darkred}{Task Description}}: Analyze the emotions of characters in the video. Implicit emotions usually need to be analyzed indirectly through the environment, style, etc.\\
\textbf{\textcolor{darkred}{Example Question}}: 

\begin{tcolorbox}[size=title,opacityfill=0.05,breakable]

\textbf{Question}: What kind of emotional atmosphere does the stage lighting create?
        
        \textbf{options}: 
        
            (A) Solemn and sorrowful atmosphere.
            
            (B) Neutral and unemotional atmosphere.
            
            (C) Intense and dramatic atmosphere.
            
            (D) Joyful and festive atmosphere.
            
            (E) Sadness and loss.
            
            (F) Confident and empowering atmosphere.
            
            (G) Chaotic and confusing atmosphere.
            
            (H) Calm and serene atmosphere.
            
            (I) Playful and whimsical atmosphere.
            
            (J) Romantic and loving atmosphere.
        
\textbf{CorrectAnswer}: (E)

\textbf{Video}: Stages of Grief- AVANTGARDE SHOW 2023
 - \href{https://www.youtube.com/watch?v=IZio9bdkzTI}{video\_url} 
\end{tcolorbox}

\subsection*{\textcolor{darkblue}{III.3. Meta-emotion}}
\textbf{\textcolor{darkred}{Task Description}}: This part refers to the high-level emotions in the video, such as the emotions expressed by the author through the video, and the emotions expressed by the entire video.\\
\textbf{\textcolor{darkred}{Example Question}}: 

\begin{tcolorbox}[size=title,opacityfill=0.05,breakable]

\textbf{Question}: Summarize the meaning of this short film in one word.

\textbf{Options}: [
            
            (A) Creation
            
            (B) Transformation
            
            (C) Stress
            
            (D) Mutation
            
            (E) Metamorphosis            
            
            (F) Growth
            
            (G) Rebirth"
            
            (H) Destruction
            
            (I) Erosion
            
            (J) Development
            
            (K) Isolation
            
            (L) Conversion
        
        \textbf{CorrectAnswer}: (C)

\textbf{Stress - Shortfilm}
 - \href{https://www.youtube.com/watch?v=WUn_SqZPN0s}{video\_url} 
\end{tcolorbox}

\subsection*{\textcolor{darkblue}{III.4. Audience Emotion}}
\textbf{\textcolor{darkred}{Task Description}}: Analyze the emotions that viewers are most likely to feel after watching the video. This is more advanced and relatively easy for humans to sense. Including the perception of humor.\\
\textbf{\textcolor{darkred}{Example Question}}: 

\begin{tcolorbox}[size=title,opacityfill=0.05,breakable]

\textbf{Question}: What are the reasons for the high number of views on this video?

\textbf{Options}: 

            (A) The video features a well-known celebrity who has a large fan base, drawing a lot of attention.
            
            (B) The dance style is extremely unique and has never been seen before, 
            sparking curiosity.
            
            (C) People are under a lot of stress and need videos that can help them unwind.
            
            (D) The background music is a popular hit song that many people recognize and enjoy.
            
            (E) The video was released during a major holiday season when people are more likely to watch videos.
            
            (F) The choreography is incredibly complex and impressive, showcasing the dancers' skills.
            
            (G) The video has a strong and inspiring message that resonates with a wide audience.
            
            (H) The video was featured on a popular TV show or news segment, driving more views.
            
            (I) The video was shared by a large number of dance schools and communities, spreading its reach.
            
            (J) The video was part of a viral challenge that encouraged people to share it.
            
            (K) The video has high-quality production values that make it stand out from other content.
        
        \textbf{CorrectAnswer}: "(C)

\textbf{Satisfying and Relaxing Kinetic Sand ASMR shorts}
 - \href{https://www.youtube.com/shorts/Cv19w_2-rMc}{video\_url} 
\end{tcolorbox}

\subsection*{\textcolor{darkblue}{IV. Comment Matching (CM)}}
\label{Appendix: CM}
\subsection*{\textcolor{darkblue}{IV.1. Humorous}}
\textbf{\textcolor{darkred}{Task Description}}: The video will spark laughter because of certain comments, making the audience feel funny, testing whether the model can match it correctly.\\
\textbf{\textcolor{darkred}{Example Question}}: 

\begin{tcolorbox}[size=title,opacityfill=0.05,breakable]

\textbf{Question}: Based on this video, which of the following comments is likely to make people laugh?
        
        \textbf{Options}: 
            
            (A) Did he just audition for a water ballet?
            
            (B) How many fish does it take to catch a man?
            
            (C) Is there a Walmart beneath the river?

            (D) The fish are holding a grudge, watch out!
            
            (E) Now that's what I call a splash of creativity.
            
            (F) I came for the fishing tips and stayed for the synchronized swimming.
            
            (G) That water has more personality than my neighbor!
            
            (H) I'm starting to think he's part fish.
            
            (I) I think the fish caught him instead.
            
            (J) That's definitely a land fish champion.
            
            (K) That fish will never trust humans again.
        
        \textbf{CorrectAnswer}: "(C)",

\textbf{He DI Lao}
 - \href{https://www.xiaohongshu.com/explore/67b0a11d000000001801802f?xsec_token=ABHIYt1VSm6MIosMt6y54hlDKDq2MTJTzjq4HMhAe1Ut8=&xsec_source=pc_search&source=unknown}{video\_url} 
\end{tcolorbox}

\subsection*{\textcolor{darkblue}{IV.2. Thought-provoking}}
\textbf{\textcolor{darkred}{Task Description}}: Some comments under the video will enhance people's thinking and test whether the model can accurately understand.\\
\textbf{\textcolor{darkred}{Example Question}}: 

\begin{tcolorbox}[size=title,opacityfill=0.05,breakable]

\textbf{Question}: Which of the following statements can better explain the social reality expressed in this animation?

\textbf{Options}: 

            (A) The animation showcases an idealized view of advancement within a corporate ladder.
            
            (B) The depiction highlights the dehumanization and mechanization of individuals in a powerful social system.
            
            (C) It portrays the joy of discovering one's true passions through societal pressures.
            
            (D) The scenes show a man achieving happiness through daily routine.
            
            (E) It represents personal ambition and the drive for success in individual careers.
            
            (F) The animation indicates the triumph of an individual's spirit in the face of adversity.
            
            (G) It reflects the disintegration of traditional family roles.
            
            (H) The animation shows the importance of family support in work-life balance.
            
            (I) It emphasizes the challenge of maintaining personal identity in urban settings.
            
            (J) We are all working for others without realizing it due to our own needs.
            
            (K) The animation illustrates the struggle with contemporary health issues.
        
        \textbf{CorrectAnswer}: (J)

\textbf{EL EMPLEO}
 - \href{https://www.youtube.com/watch?v=cxUuU1jwMgM}{video\_url} 
\end{tcolorbox}

\subsection*{\textcolor{darkblue}{IV.3. Trending}}
\textbf{\textcolor{darkred}{Task Description}}: It is relatively difficult to test whether the model can accurately infer and analyze the most popular comments under the video.\\
\textbf{\textcolor{darkred}{Example Question}}: 

\begin{tcolorbox}[size=title,opacityfill=0.05,breakable]

\textbf{Question}: Which of the following comments best summarizes the content conveyed by this film?

\textbf{Options}: 

            (A) Material possessions define one's value.
            
            (B) Selfless acts lead to rewards that surpass material wealth.
            
            (C) Loneliness is a desirable state.
            
            (D) Personal gains are the ultimate goal of helping others.
            
            (E) Isolation is the path to personal growth.
            
            (F) True happiness is found through wealth accumulation.
            
            (G) Success comes from competitive behavior.
            
            (H) Sharing leads to financial prosperity.
            
            (I) He receives what money can't buy.
            
            (J) Adversity breeds stronger individuals.
        
        \textbf{CorrectAnswer}: (I)

\textbf{Unsung Hero}
 - \href{https://www.youtube.com/watch?v=uaWA2GbcnJU}{video\_url} 
\end{tcolorbox}

\subsection*{\textcolor{darkblue}{V. Implicit Symbol (IS)}}
\label{Appendix: IS}
\subsection*{\textcolor{darkblue}{V.1. Cultural Symbols}}
\textbf{\textcolor{darkred}{Task Description}}: Test whether the model can infer and analyze the cultural characteristics hidden under the surface visual elements of the video (such as nationality, festivals, customs, religion, etc.).\\
\textbf{\textcolor{darkred}{Example Question}}: 

\begin{tcolorbox}[size=title,opacityfill=0.05,breakable]

\textbf{Question}: The plaque inscribed with “Dominating Three Continents” that appears in the video is most likely to be found in the architecture of which of the following religions?

\textbf{Options}: 

            (A) Taoism
            
            (B) Shinto
            
            (C) Sikhism
            
            (D) Judaism
            
            (E) Islam
            
            (F) Christianity
            
            (G) Buddhism
            
            (H) Hinduism
            
            (I) Jainism
            
            (J) Zoroastrianism
        
        \textbf{CorrectAnswer}: (G)

\textbf{[4K] Hangzhou 2024 in the misty rain | West Lake, Lingyin Temple, Night walking in Hefang Street}
 - \href{https://www.youtube.com/watch?v=61MR4EqU9Vg&t=3317s}{video\_url} 
\end{tcolorbox}

\subsection*{\textcolor{darkblue}{V.2. Art Symbols}}
\textbf{\textcolor{darkred}{Task Description}}: Test whether the model can infer and analyze the art-related characteristics hidden under the surface visual elements of the video (such as dance style, artistic skills, imitation, etc.).\\
\textbf{\textcolor{darkred}{Example Question}}: 

\begin{tcolorbox}[size=title,opacityfill=0.05,breakable]

\textbf{Question}: What is the shadow that appears in our view at 1:40 imitating?
        
        \textbf{Options}: 
        
            (A) The shadow is imitating a pole dancer.
            
            (B) The shadow is imitating a person washing a dog.
            
            (C) The shadow is imitating a person brushing their hair.
            
            (D) The shadow is imitating someone playing a violin.
            
            (E) The shadow is imitating two people engaged in a conversation.
            
            (F) The shadow is imitating someone painting a wall.
            
            (G) The shadow is imitating a person feeding a horse.
            
            (H) The shadow is imitating a person washing their car.
            
            (I) The shadow is imitating a dog barking at a person.
        
            (J) The shadow is imitating someone performing a magic trick.
            
            (K) The shadow is imitating a person holding an umbrella.
            
            (L) The shadow is imitating someone walking a large dog.

            \textbf{CorrectAnswer}: (A)

\textbf{LEAKED! Hilarious Shadow Puppets - AGT 2023 Early Release}
 - \href{https://www.youtube.com/watch?v=0JfGZyGlIZY&t=5s}{video\_url} 
\end{tcolorbox}

\subsection*{\textcolor{darkblue}{V.3. Other Symbols}}
\textbf{\textcolor{darkred}{Task Description}}: Test whether the model can infer and analyze other special symbols (such as commercial advertisements, etc.) hidden under the surface visual elements of the video.\\
\textbf{\textcolor{darkred}{Example Question}}: 

\begin{tcolorbox}[size=title,opacityfill=0.05,breakable]

\textbf{Question}: "What do you think the chimpanzee that appears multiple times in the film symbolizes?",

\textbf{Options}: 

            (A) The chimpanzee symbolizes chaos and disruption in everyday life.
            (B) The chimpanzee symbolizes a childhood fear.
            (C) The chimpanzee symbolizes technology invading personal space.
            (D) The chimpanzee symbolizes the unpredictability of fate.
            (E) The chimpanzee symbolizes a glue company.
            (F) The chimpanzee symbolizes lost opportunities.
            (G) The chimpanzee symbolizes an obsession with social status.
            (H) The chimpanzee symbolizes environmental degradation.
            (I) The chimpanzee symbolizes the desire for freedom.
            (J) The chimpanzee symbolizes misunderstanding between people.
            (K) The chimpanzee symbolizes reliability and trust in friendships.
        
        \textbf{CorrectAnswer}: (E)

\textbf{All Gorilla glue ads}
 - \href{https://www.youtube.com/watch?v=ieQq9pGGs-I}{video\_url} 
\end{tcolorbox}

\subsection{Explicit Reasoning Tasks}

\subsection*{\textcolor{darkblue}{I. Causal Reasoning (CAR)}}
\subsection*{\textcolor{darkblue}{I.1. Forward Reasoning}}
\textbf{\textcolor{darkred}{Task Description}}: Forward reasoning can also be understood as the prediction of future events, including prediction of outcomes, prediction of content that has not yet appeared, etc.\\
\textbf{\textcolor{darkred}{Example Question}}: 

\begin{tcolorbox}[size=title,opacityfill=0.05,breakable]

\textbf{Question}: What is the speculated ending of the film?
        
        \textbf{Options}: 
        
            (A) The movie concludes with an unexpected twist where the flowers reveal a hidden secret.
            
            (B) The ending is a cliffhanger, leaving the audience uncertain about the characters' fate.
            
            (C) Her boyfriend passed away due to illness, leaving the girl devastated with grief.
            
            (D) The film wraps up with a joyous family reunion.
            
            (E) The film ends with a dramatic breakup as one character leaves with a heavy heart.
            
            (F) The movie concludes with a comedic mishap involving the flowers.
            
            (G) The ending shows a tragic farewell as one character moves to a new city.
            
            (H) The film ends with the revelation of a long-lost sibling.
            
            (I) The story concludes with the characters embarking on a spontaneous road trip.
            
            (J) The film ends on a melancholic note, reflecting on lost opportunities.
            
            (K) The video closes with a heartwarming reconciliation between the main characters after exchanging heartfelt notes and gestures.
        
        \textbf{CorrectAnswer}: (C)
        
\textbf{For Milo - AWARD WINNING 1 Minute Short film (2020)}
 - \href{https://www.youtube.com/watch?v=9Tq71PiDJDk}{video\_url} 
\end{tcolorbox}

\subsection*{\textcolor{darkblue}{I.2. Backward Reasoning}}
\textbf{\textcolor{darkred}{Task Description}}: Backward reasoning means finding the cause from the effect and inferring the reason why an event occurred.\\
\textbf{\textcolor{darkred}{Example Question}}: 

\begin{tcolorbox}[size=title,opacityfill=0.05,breakable]

\textbf{Question}: Why was the elderly black man warned by security at the beginning of the film?

        \textbf{Options}: 
        
            (A) Mobile phones are not allowed for recording during magic shows.
            
            (B) He was trying to sell unauthorized merchandise.
            
            (C) He was recognized as a local celebrity causing disruptions.
            
            (D) He was accused of stealing a bicycle.
            
            (E) He was creating loud music disturbing the peace.
            
            (F) He was believed to have lost his entrance ticket.
            
            (G) He was inadvertently blocking the pathway.
            
            (H) He was associated with another person causing trouble nearby.
            
            (I) He was engaged in card tricks that security found suspicious.
            
            (J) He was loitering without a purpose.
        
        \textbf{CorrectAnswer}: (A)
        
\textbf{Now You See Me Official Opening Scene (2013) - Mark Ruffalo, Morgan Freeman Movie HD}
 - \href{https://www.youtube.com/watch?v=u_diRgwPCS8&t=2s}{video\_url} 
\end{tcolorbox}

\subsection*{\textcolor{darkblue}{II. Sequential Structure Reasoning (SSR)}}
\subsection*{\textcolor{darkblue}{II.1. Narrative Structure}}
\textbf{\textcolor{darkred}{Task Description}}: Reasoning and analyzing the narrative order of the entire video, including the editing order, such as sequential, flashback, and interpolation.\\
\textbf{\textcolor{darkred}{Example Question}}: 
\begin{tcolorbox}[size=title,opacityfill=0.05,breakable]

\textbf{Question}: What kind of narrative sequence does the film employ?
        
        \textbf{Options}: 
        
            (A) non-linear flashback sequence, where events are shown out of chronological order, often revealing backstory
            
            (B) parallel overlapping sequences, showing multiple storylines happening simultaneously with some overlap
            
            (C) cyclical narrative structure, repeating events or themes in a circular pattern
            
            (D) linear narrative sequence, following a straightforward progression from beginning to end
            
            (E) random jumps in the timeline, moving unpredictably between different points in time
            
            (F) interwoven thematic structure, weaving together different themes and ideas throughout the story
            
            (G) reverse chronological order, starting with the end and moving backwards in time
            
            (H) fragmented narrative, presenting the story in disjointed or broken segments
            
            (I) begins with a flashback and then proceeds in chronological order
            
            (J) episodic progression, advancing the story through a series of distinct episodes or chapters
            
            (K) multi-perspective narrative, telling the story from multiple characters' points of view
        
        \textbf{CorrectAnswer}: (I)
        
\textbf{Identity SHORT FILM (Award Winning Inspirational Short)}
 - \href{https://www.youtube.com/watch?v=ikGVWEvUzNM}{video\_url} 
\end{tcolorbox}

\subsection*{\textcolor{darkblue}{II.2. Core Connecting Elements}}
\textbf{\textcolor{darkred}{Task Description}}: Videos with this type of question usually have a key connecting element that runs through the entire video. It is carefully designed by the producer and tests the model's inductive reasoning of the visual information of the entire video.\\
\textbf{\textcolor{darkred}{Example Question}}: 
\begin{tcolorbox}[size=title,opacityfill=0.05,breakable]

\textbf{Question}: What is the recurring element in the video, summarized in one word?
        
        \textbf{Options}: 
        
            (A) Pareidolia
            
            (B) Smile
            
            (C) Alarm
            
            (D) Work
            
            (E) Mirror
            
            (F) Mundane
            
            (G) Routine
            
            (H) Suit
            
            (I) Coffee
            
            (J) Sleep
            
            (K) Bedroom
            
            (L) Portrait
        
        \textbf{CorrectAnswer}: (B)
        
\textbf{PAREIDOLIA - 1 Minute Short Film | Award Winning}
 - \href{https://www.youtube.com/watch?v=JLmOkEEC9SQ}{video\_url} 
\end{tcolorbox}

\subsection*{\textcolor{darkblue}{II.3. Inference on Editing Techniques}}
\textbf{\textcolor{darkred}{Task Description}}: These tasks evaluate the models’ deep analysis and multimodal reasoning about video editing strategies.\\
\textbf{\textcolor{darkred}{Example Question}}: 
\begin{tcolorbox}[size=title,opacityfill=0.05,breakable]
\textbf{Question}: "Please guess how many videos were needed to record the moment the man punched the punctured water ball at the beginning of the video?",

\textbf{Options}: 
            
            (A) At least two separate takes would be needed.
            
            (B) At least one single take is needed.
            
            (C) Three separate takes are needed.
            
            (D) Four separate takes are needed.
            
            (E) Each scene can be captured in a single continuous take.
            
            (F) Five separate takes are needed.
            
            (G) Six separate takes are needed.
            
            (H) Eight separate takes are needed.
        
            (I) Ten separate takes are required.
            
            (J) Twenty separate takes are necessary.
            
            (K) At least ten separate takes are needed.
        
        \textbf{CorrectAnswer}: (C)
        
\textbf{Playing With Time}
 - \href{https://www.youtube.com/watch?v=gooWdc6kb80}{video\_url} 

\textbf{Note:} The reasoning and analysis process of this question can refer to this \href{https://www.bilibili.com/video/BV1c3ZwY4EqE?spm_id_from=333.788.recommend_more_video.-1&vd_source=e2638f46408a99009fc4299e944cf139}{disassembly video} .
\end{tcolorbox}

\subsection*{\textcolor{darkblue}{II.4. Hallucination}}
\textbf{\textcolor{darkred}{Task Description}}: Evaluate whether the model perceives various types of hallucinations when perceiving video content.\\
\textbf{\textcolor{darkred}{Example Question}}: 
\begin{tcolorbox}[size=title,opacityfill=0.05,breakable]
\textbf{Question}: How many dancers are there in the video?
        
        \textbf{Options}: 
        
            (A) 0
            
            (B) 1
            
            (C) 2
            
            (D) 3
            
            (E) 4
            
            (F) 5
            
            (G) 6
            
            (H) 7
            
            (I) 8
            
            (J) 9
            
            (K) options before are all false
        
        \textbf{CorrectAnswer}: (B)
        
\textbf{Rat dance with falling body parts}
 - \href{https://www.youtube.com/shorts/CzIJjj-B_JY}{video\_url} 
\end{tcolorbox}

\subsection*{\textcolor{darkblue}{III. Counterintuitive Reasoning (CIR)}}
\subsection*{\textcolor{darkblue}{III.1. Magic Deconstruction or Special Effects Editing}}
\textbf{\textcolor{darkred}{Task Description}}: This type of video usually creates some impossible magical effects, but some are magic tricks, and some are editing and special effects, which require deeply reasoning.\\
\textbf{\textcolor{darkred}{Example Question}}: 
\begin{tcolorbox}[size=title,opacityfill=0.05,breakable]
\textbf{Question}: Starting at 4:35, how did the man achieve this magical effect in the magic trick?
        
        \textbf{Options}: 
            (A) Sleight of hand technique with a hidden ring, using dexterity to conceal and reveal the ring.
            
            (B) Utilizing a mirror to confuse the audience, creating optical illusions through reflection.
            
            (C) A distraction technique with a smoke bomb, diverting attention with a sudden burst of smoke.
            
            (D) A special ring that retracts into a fake thumb, using a concealed mechanism to make the ring disappear.
            
            (E) Using a magnet hidden in the sleeve, manipulating objects with magnetic force.
            
            (F) A camera trick with video editing, altering footage to create the illusion of magic.
            
            (G) Sleight of hand technique with a hidden string, using a concealed thread to control objects.
            
            (H) The bottle inside the paper bag had already been altered to leave only the outer plastic skin.
            
            (I) Employing a twin assistant to swap the ring, using a look-alike to deceive the audience.
            
            (J) The use of an invisible thread, employing a nearly undetectable line to move objects.
            
            (K) A sound cue to mislead the audience's attention, using noise to distract from the real action.
        
        \textbf{CorrectAnswer}: (H)
        
\textbf{Level 1 to 100 Magic Tricks Anyone Can Do}
 - \href{https://www.youtube.com/watch?v=eGewFYQhJEQ&t=54s}{video\_url} 
\end{tcolorbox}

\subsection*{\textcolor{darkblue}{III.2. Artistic Techniques}}
\textbf{\textcolor{darkred}{Task Description}}: This type of video usually creates some impossible scenes, but it is usually an artistic expression deliberately designed by the author.\\
\textbf{\textcolor{darkred}{Example Question}}: 
\begin{tcolorbox}[size=title,opacityfill=0.05,breakable]
\textbf{Question}: Why is the shadow on the boy's face illuminated by sunlight at 1:06?
        
        \textbf{Options}: 
            (A) Because the boy moves to a position where a strong light source is directly above him, not related to the girl.
            
            (B) It's just a coincidence that the angle of the sun changes suddenly at that moment, and has nothing to do with any special meaning.
            
            (C) The sunlight illuminates the shadow because the cameraman adjusts the lighting equipment to create a better visual effect.
            
            (D) The girl's appearance brings good luck, and the sunlight representing good fortune clears away the gloom of bad luck in his world.
            
            (E) This is because the boy has walked into a neighborhood with better weather and climate.
            
            (F) The sunlight lights up the shadow because there is a hidden light - emitting device in the scene that is turned on at 1:06.
            
            (G) It's a result of the special lens filter used during filming, which makes the shadow on the boy's face appear to be lit by sunlight.
            
            (H) Because the boy didn't get hurt after falling and his mood improved, the sunlight is used to represent his improved mood.
        
        \textbf{CorrectAnswer}: (D)
        
\textbf{CGI Animated Short Film HD "Jinxy Jenkins \& Lucky Lou" by Mike Bidinger \& Michelle Kwon | CGMeetup
}
 - \href{https://www.youtube.com/watch?v=OuJ4BBQ0nhc}{video\_url} 
\end{tcolorbox}

\subsection*{\textcolor{darkblue}{III.3. Humor and Exaggeration}}
\textbf{\textcolor{darkred}{Task Description}}: A common technique in humorous videos is to use exaggerated expressions that seem unreasonable, but there are some clues to understand the meaning. This type of question tests the model's ability to reason about exaggerations and unusual techniques.\\
\textbf{\textcolor{darkred}{Example Question}}: 
\begin{tcolorbox}[size=title,opacityfill=0.05,breakable]
\textbf{Question}: Why does the first half of the scene look sunny but also show rain?
        
        \textbf{Options}: 
            
            (A) It is a sunshower, when rain falls while the sun is shining.
            
            (B) The character is dreaming of being both wet and warm.
            
            (C) There are rainclouds directly above while sunlight comes from the side.
            
            (D) It is snow instead of rain, reflecting the sunlight.
            
            (E) The effect is caused by morning fog and light refraction.
            
            (F) It's a visual illusion caused by mist.
            
            (G) The character moved to a different location quickly.
            
            (H) A rainbow is forming which intensifies the sunlight.
            
            (I) Dew drops from trees reflect sunlight.
            
            (J) There are two unrelated weather animations merged together.
            
            (K) The man wet the bed, which caused the presence of water in his dream.
        
        \textbf{CorrectAnswer}: (K)
        
\textbf{It now makes sense}
 - \href{https://www.youtube.com/shorts/xPX9wOM3TnI}{video\_url} 
\end{tcolorbox}

\subsection*{\textcolor{darkblue}{IV. Cross-modal Transfer Reasoning (CTR)}}
Evaluate the ability to transfer reasoning from video to text, audio, video or image (for example, video-to-text: the theme of a video may have the same meaning as a famous quote)
\textbf{\textcolor{darkred}{Task Description}}: Evaluate the ability to transfer reasoning from video to text (for example, the theme of a video may have the same meaning as a famous quote)\\
\textbf{\textcolor{darkred}{Example Question}}: 
\begin{tcolorbox}[size=title,opacityfill=0.05,breakable]
\textbf{Question}: Which of the following proverbs best explains the theme of this short film?
        
        \textbf{Options}: 
        
            (A) When one door closes, another opens.
            
            (B) Opportunity knocks only once.
            
            (C) Time heals all wounds.
            
            (D) The early bird catches the worm.
            
            (E) Never judge a book by its cover.
            
            (F) All that glitters is not gold.
            
            (G) The grass is always greener on the other side.
            
            (H) Actions speak louder than words.
            
            (I) A stitch in time saves nine.
            
            (J) Beauty is in the eye of the beholder.
            
            (K) Absence makes the heart grow fonder.
            
            (L) A penny saved is a penny earned.
        
        \textbf{CorrectAnswer}: (E)
        
\textbf{Video}: Award Winning SHORT FILMS Don't Judge | BATTI Hindi Heart Touching Short Movies | Content Ka Keeda
 - \href{https://www.youtube.com/watch?v=3-PXNnaatx4}{video\_url} 
\end{tcolorbox}

\subsection*{\textcolor{darkblue}{V. Video Type and Intent (VTI)}}
\subsection*{\textcolor{darkblue}{V.1. Video Type}}
\textbf{\textcolor{darkred}{Task Description}}: Evaluate the model's ability to analyze video types, such as commercials, science fiction films, comedies, etc.\\
\textbf{\textcolor{darkred}{Example Question}}: 
\begin{tcolorbox}[size=title,opacityfill=0.05,breakable]
\textbf{Question}: What type of video is this most likely to be?
        
        \textbf{Options}: 
        
            (A) A documentary about airplane technology
            
            (B) Advertisement for an ice-cream
            
            (C) A drama set on an airplane
            
            (D) A comedy film featuring an airline
            
            (E) An in-flight safety demonstration video
            
            (F) A travel vlog featuring aerial views
            
            (G) A science fiction movie on a spaceship
            
            (H) This is an advertisement.
            
            (I) A video tour of an airplane factory
            
            (J) A virtual reality experience of flying
            
            (K) A news segment on turbulence incidents
        
        \textbf{CorrectAnswer}: (H)
        
\textbf{Leo Messi vs Kobe Bryant - Legends on Board - Turkish Airlines}
 - \href{https://www.youtube.com/watch?v=ApPkdTNbcY8}{video\_url} 
\end{tcolorbox}

\subsection*{\textcolor{darkblue}{V.2. Video Intent}}
\label{Appendix: VTI}
\textbf{\textcolor{darkred}{Task Description}}: Reasoning and analyzing the purpose and production intention of the video (e.g. what kind of product performance is promoted in a commercial advertisement, etc.)\\
\textbf{\textcolor{darkred}{Example Question}}: 
\begin{tcolorbox}[size=title,opacityfill=0.05,breakable]
\textbf{Question}: Which year do you think this video was most likely released?
        
        \textbf{Options}:
        
            (A) 2018
            
            (B) 2017
            
            (C) 2016
            
            (D) 2015
            
            (E) 2023
            
            (F) 2019
            
            (G) 2023
            
            (H) 2020
            
            (I) 2014
            
            (J) 2013
        
        \textbf{CorrectAnswer}: (H)
        
\textbf{Lockdown | One Minute Short Film Challenge | Film Riot}
 - \href{https://www.youtube.com/watch?v=Jw-FcDdxSU8}{video\_url} 
\end{tcolorbox}

\section{Evaluation Details}
\label{sec: exp details}

\subsection{Baselines}
\label{appendix:baselines}
The baselines include closed-source models: (1) GPT series: GPT-4o~\citep{GPT-4o}, GPT-4o-mini~\citep{GPT-4o-mini}, GPT-4.1~\citep{openai2025gpt41}, and GPT-5~\citep{gpt5}; (2) Gemini series: Gemini-2.0-flash, Gemini-2.0-flash-thinking-01-21~\citep{reid2024gemini}, and Gemini-2.5 (flash/pro)~\citep{google2025gemini2.5}; (3) Claude-3-5-Sonnet-20241022~\citep{claude-3-5}; (4) o3, o4-mini~\citep{o3}; open-source models: (1) Qwen series: Qwen2.5-VL (7B/72B-Instruct)~\citep{qwen25}; (2) Gemma series: Gemma-3 (12B/27B)~\citep{gemma3}; (3) InternVL series: Intern3-VL (8B/38B)~\citep{zhu2025internvl3}; (4) LLava series: LLava-Onevision-7B~\citep{li2024llava}, Video-LLava-7B~\citep{lin2023video}; (5) Phi-4-multimodal-Instruct~\citep{phi4}; (6) Other video models: Cogvlm2-video-llama3-chat~\citep{hong2024cogvlm2}, NVILA-8B-Video~\citep{liu2024nvila}. All local experiments are conducted on 4×A100 80GB GPUs.

\subsection{Frame Selection}
We followed the official configurations of models that support multi-image input, as well as settings in previous works~\citep{fu2024video, wang2024lvbench}, to define the number of input frames for each model. Specifically, we fixed the number of frames per model and sampled them evenly across the video duration. We sampled 8 frames for LLaVA-OneVision, Video-LLaVA, and NVILA-8B-Video, Phi-4-multimodal-instruct; 16 frames were sampled for Qwen2.5-VL-7B, Qwen2.5-Omni-7B, CogVLM2-Video-LLaMA3-Chat, InternVL-8B, Gemma-3-it-12B and Gemini-2.0-Flash-Thinking; 32 frames were sampled for Qwen2.5-VL-72B, InternVL-38B, Gemma-3-it-27B, GPT-4o-Mini, o4-mini, o3, GPT-5, Gemini-2.5-Flash-preview and Claude-3.5-Sonnet; 300 frames for GPT-4o.
Exceptionally, since Gemini-2.0-Flash supports long video and multimodal context inputs, we sampled one frame per second across each video, with a maximum cap of 512 frames to ensure API stability. For Gemini-2.5-pro, we also tested the official setting of 1fps sampling frequency. Additionally, to enable a fair comparison with Gemini-2.0-Flash-Thinking, we also tested a version of Gemini-2.0-Flash with 16 frames.


\section{Further Results}

\subsection{Impact of Video Length on Model Performance}
\label{sec:appendix_video_length}

To provide a more comprehensive analysis, we evaluated model performance on videos of varying lengths. Following the categorization standards used in prior work such as Video-MME~\citep{fu2024video} and based on the video length distribution of MMR-V, we divided the dataset into three groups: \textbf{Short} ($<2$ minutes), \textbf{Medium} ($2$--$6$ minutes), and \textbf{Long} ($>6$ minutes). These categories comprise 329, 593, and 335 tasks, respectively. The detailed evaluation results are presented in Table~\ref{tab:video_length_perf}. We observe a general trend where the performance of most models decreases as the video length increases. However, it is worth noting that this performance gap becomes \textbf{less pronounced for models equipped with long-context capabilities}. For instance, \textit{Gemini-2.0-Flash} (processing 512 frames) and \textit{Gemini-2.5-Pro} (using 1 fps sampling) demonstrate remarkable stability across all duration categories.

\begin{table*}[t]
\centering
\normalsize
\renewcommand{\arraystretch}{1.2} 
\setlength{\tabcolsep}{3.8pt}
\caption{Evaluation results (\%) across different video length. \textbf{Bold} and \underline{underlined} values indicate the best performance among proprietary and open-source models, respectively.}
\resizebox{\linewidth}{!}{%
\begin{tabular}{p{4.5cm}|m{2.2cm}|m{2.2cm}|m{2.2cm}|m{2.2cm}}
\hline
 &  \multicolumn{1}{c|}{} & \multicolumn{3}{c}{\textbf{Video Length}} \\
\cline{3-5}
\textbf{Model} & \multicolumn{1}{c|}{\textbf{Overall}} & \multicolumn{1}{c|}{\textbf{Short Video}} & \multicolumn{1}{c|}{\textbf{Medium Video}} & \multicolumn{1}{c}{\textbf{Long Video}} \\ 
\hline
\multicolumn{5}{c}{\emph{Open-source models}} \\
\textrm{LLaVA-Video} & 18.4 & 20.5 & 16.7 & 10.4 \\
\textrm{NVILA-8B-Video} & 25.5 & 27.1 & 25.9 & 22.6 \\
\textrm{Phi-4-multimodal-instruct} & 26.7 & 30.1 & 24.5 & 27.2 \\
\textrm{Cogvlm2-video-llama3} & 25.6 & 27.4 & 25.9 & 22.6 \\
\textrm{Qwen2.5-VL-7B} & 30.1 & 34.7 & 28.0 & 29.3 \\
\textrm{InternVL3-8B} & 33.6 & 34.9 & 33.0 & 32.8 \\
\textrm{Gemma-3-12b-it} & 34.0 & 35.7 & 33.5 & 31.1 \\
\textrm{InternVL2.5-38B} & 39.9 & \underline{43.9} & 39.1 & 38.2 \\
\textrm{Qwen2.5-VL-72B} & 39.1 & 43.8 & 40.4 & 34.5 \\
\textrm{Gemma-3-27b-it} & \underline{42.0} & 43.8 & \underline{42.0} & \underline{40.3} \\
\hline
\multicolumn{5}{c}{\emph{Proprietary models}} \\
\textrm{GPT-4o-mini-2024-07-18} & 34.8 & 39.2 & 34.1 & 31.6 \\
\textrm{Gemini-2.0-Flash (16 frames)} & 42.6 & 46.2 & 42.3 & 39.7 \\
\textrm{Claude-3-5-Sonnet-20241022} & 43.3 & 47.7 & 43.5 & 38.5 \\
\textrm{Gemini-2.0-Flash-thinking} & 45.0 & 48.9 & 45.0 & 40.9 \\
\textrm{GPT-4.1-2025-04-14} & 46.6 & 49.9 & 46.2 & 44.2 \\
\textrm{Gemini-2.0-Flash (512 frames)} & 48.8 & 48.9 & 48.6 & 49.0 \\
\textrm{Gemini-2.5-Flash} & 51.2 & 54.1 & 51.9 & 47.2 \\
\textrm{o4-mini-2025-04-16} & 52.5 & 57.1 & 53.0 & 47.2 \\
\textrm{GPT-4o-2024-11-20} & 52.8 & 57.4 & 53.1 & 48.1 \\
\textrm{o3-2025-04-16} & 59.1 & 62.9 & 59.7 & 54.3 \\
\textrm{GPT-5-2025-08-07} & 60.9 & 62.3 & 61.6 & 56.2 \\
\textrm{Gemini-2.5-pro (1fps)} & \textbf{64.3} & \textbf{64.7} & \textbf{64.1} & \textbf{63.9} \\
\hline
\end{tabular}%
}
\label{tab:video_length_perf}
\vspace{-10pt}
\end{table*}

\subsection{Benefit of Audio Input}
\label{appendix:audio}

In this section, we illustrate through several examples how incorporating audio as an additional modality provides extra reasoning cues that enhance video reasoning. In the following examples, green text highlights the model’s analysis of audio cues, which contributes to reaching the final answer.

\begin{figure}[t]
    \centering
    \includegraphics[width=\linewidth]{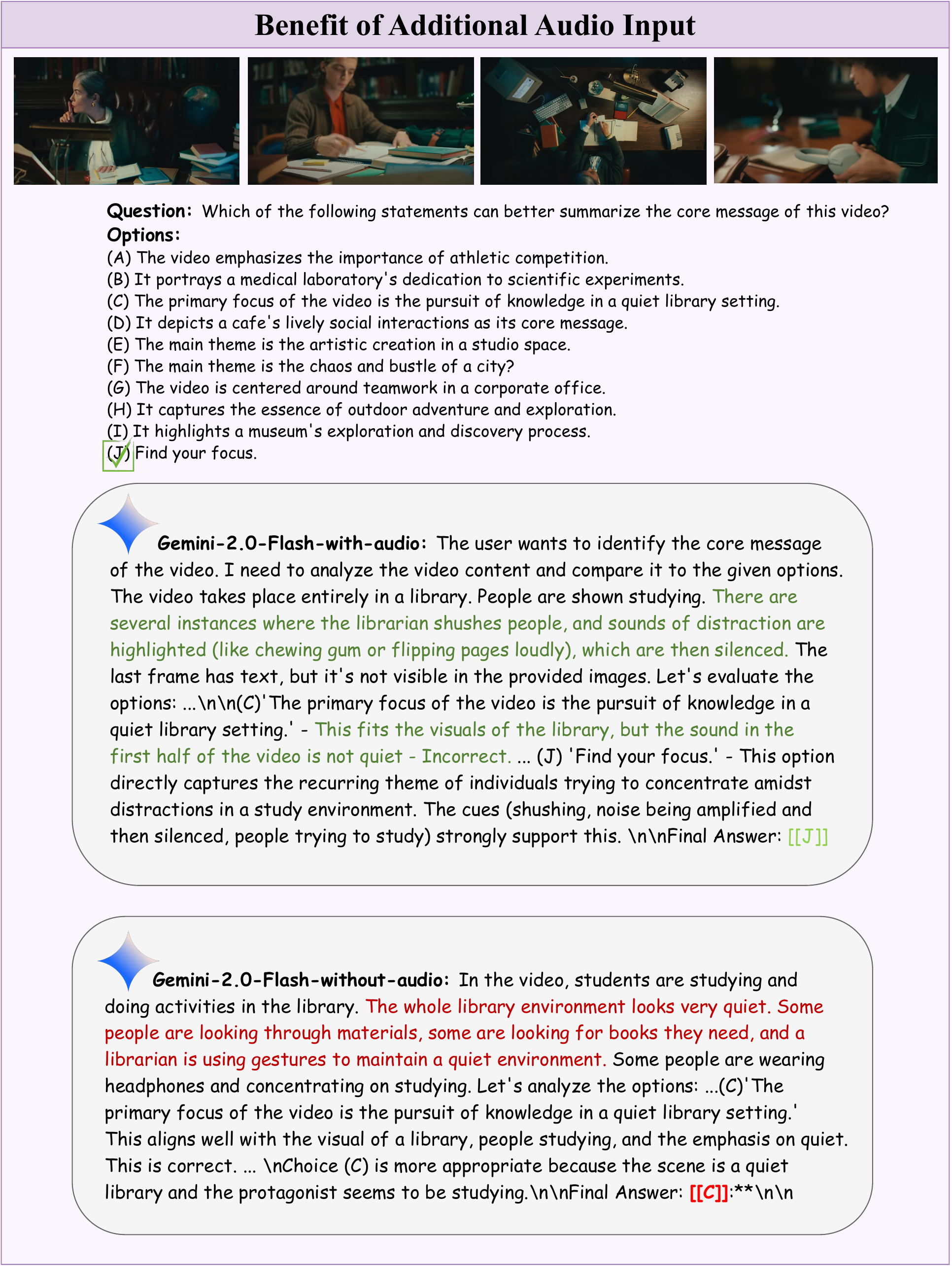}
    \caption{Benefit of Additional Audio Input. The green text highlights the analysis of audio cues.}
    \label{fig:audio impact 1}
\vspace{-1em}
\end{figure}

In the first example (Figure \ref{fig:audio impact 1}), audio reveals that the library is actually noisy. When the protagonist puts on headphones, the video turns silent. Without audio, the model might mistakenly assume the library is quiet based solely on visuals. The original video link is: \href{https://www.youtube.com/watch?v=oYmU8Av_e84}{Spec commercial - SONY}.

\begin{figure}[t]
    \centering
    \includegraphics[width=\linewidth]{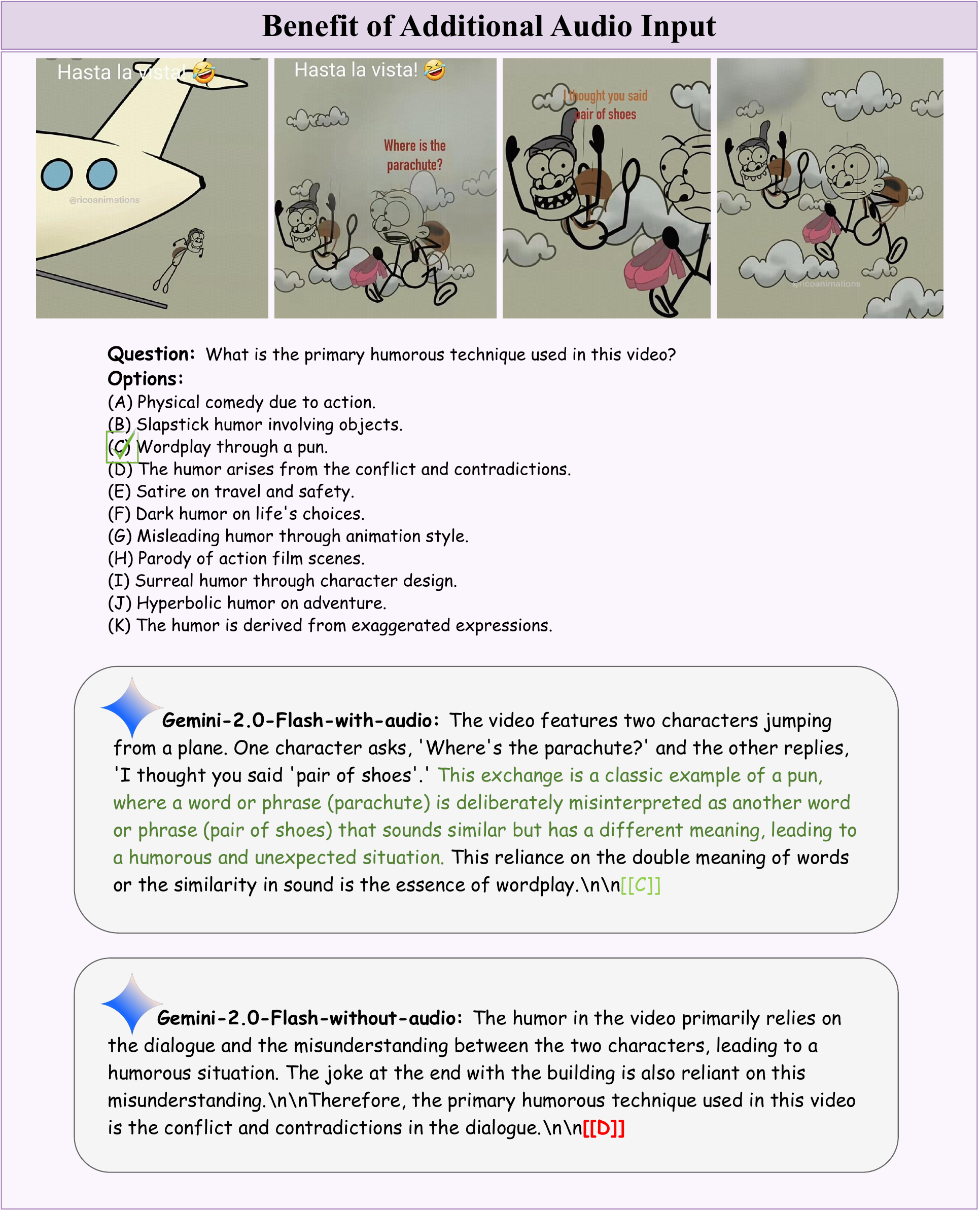}
    \caption{Benefit of Additional Audio Input. The green text highlights the analysis of audio cues.}
    \label{fig:audio impact 2}
\vspace{-1em}
\end{figure}

In the second example (Figure \ref{fig:audio impact 2}) the audio can better understand that 'parachute' and 'pair of shoes' are homophones. The original video link is: \href{https://www.youtube.com/shorts/5ubYYmPVEN8}{Hasta la vista}. 

Moreover, audio can supply cues from other perspectives. For instance, background music can convey the overall mood of a video, while magic performance videos often include voice narration.
\clearpage
\section{Details of CoT Analysis Experiment}
\label{appendix:cot_analysis}

\begin{figure}[t]
    \centering
    \includegraphics[width=\linewidth]{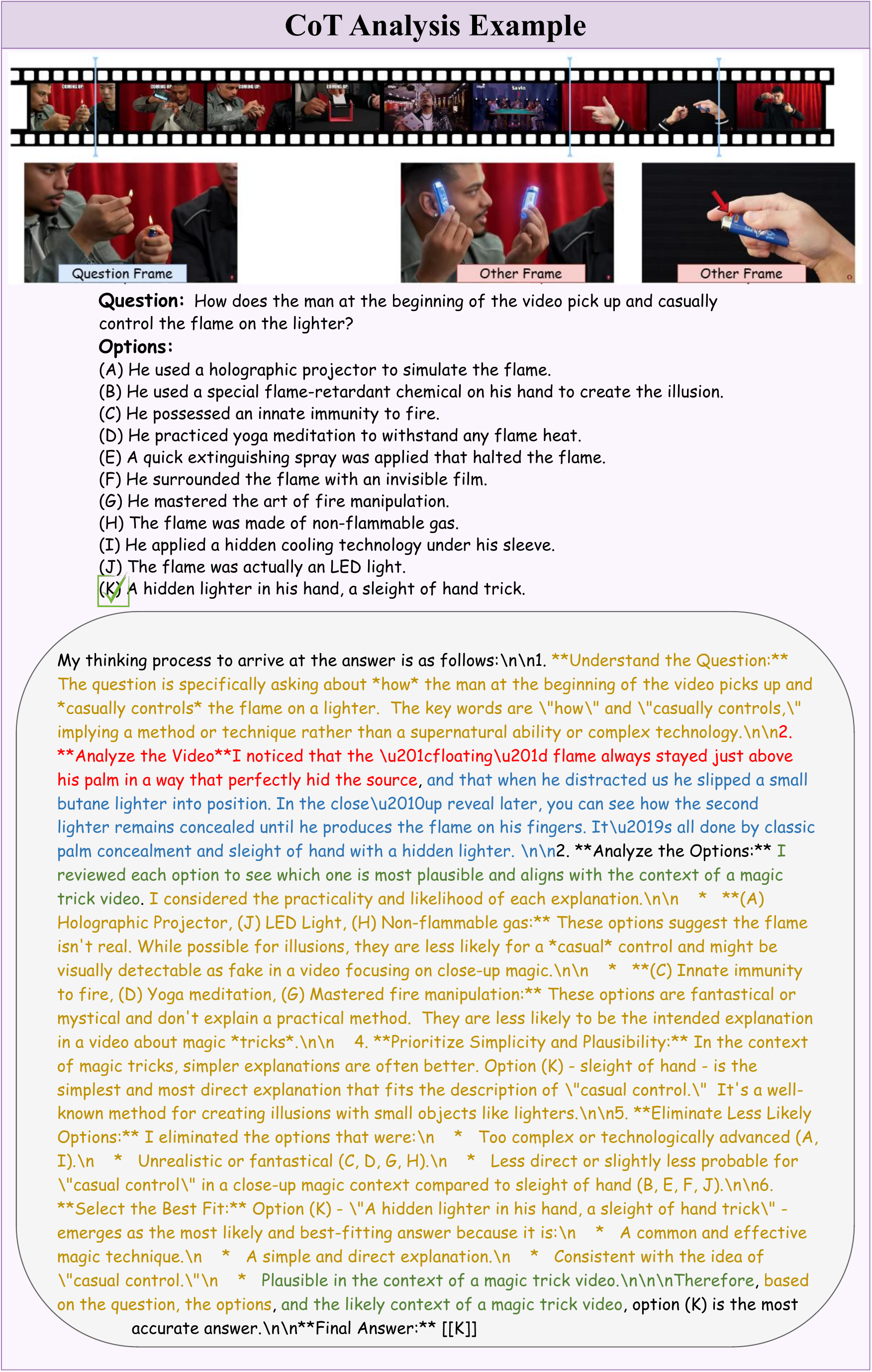}
    \caption{CoT example of experiments in Section \ref{sec:error analysis}. Yellow, red, and blue represent text, question frame, and other frame analysis respectively. Red, blue, and green all represent video analysis.}
    \label{fig:cot analysis}
\vspace{-1em}
\end{figure}

This section presents the CoT analysis experiments discussed in Section \ref{sec:error analysis}. We use a representative model CoT to illustrate 4 categories of analysis in Figure \ref{fig:cot analysis}. Specifically, Text Analysis refers to the examination of textual information such as the question and options; Video Analysis focuses on the content of the video; Question Frame targets the specific frame referenced in the question—for instance, the frame where the magician controls two flames; and Other Frame pertains to frames outside the scope of the question.
In Figure \ref{fig:cot analysis}, yellow, red, and blue represent text, question frame, and other frame analysis respectively. Red, blue, and green all represent video analysis.

\begin{table*}[h]

\centering
    \caption{    
    CoT analysis prompt.
    }
\resizebox{\linewidth}{!}{    \small
    \begin{tabular}{p{\linewidth}}
        \toprule
        \underline{\textbf{Prompt for CoT Annotation}} \\
        \\[1mm]

You will be given a model's textual reply to a video-based question along with the video frames. Your task is to determine four boolean labels for each chunk of the reply: \\

1. `other frame desc`: Does this chunk describe visual information from frames other than question frame?  \\
2. `question frame desc`: Does this chunk correctly describe visual information from the question frame specified in the question?  \\
3. `video analysis`: Does this chunk perform analysis of the video content? \\
4. `text analysis`: Does this chunk perform analysis of the text (e.g., question text, options) rather than visual content  \\

- The question frame refers to the specific frame(s) referenced by the question prompt. \\
- Other-frame descriptions are visual details not present in the question frame but from other frames. \\
- Video analysis includes describing trends, motions, or visual inference beyond plain description. \\
- Text analysis includes reasoning over question text, options, or external text context. \\

Respond strictly in JSON: \\
\{ \\
"other frame desc": true or false, \\
  "question frame desc": true or false, \\
  "video analysis": true or false, \\
  "text analysis": true or false \\
\}  \\

Question: \colorbox{mygray}{\{question\}} \\
Reply Chunk: \colorbox{mygray}{\{chunk\}} \\
Whole CoT Reply: \colorbox{mygray}{\{CoT\}} \\
    \bottomrule
    \end{tabular}}
    \label{appendix:cot_analysis_prompt}
    
\end{table*}

\clearpage
\section{Evaluation Prompt}
\label{sec:prompt}
We evaluated two settings in the main experiment, zero shot and zero shot + CoT. The prompts used are as follows.

\begin{table*}[h]

\centering
    \caption{    
    Evaluation prompt for the Zero-Shot Setting.
    }
\resizebox{1.05\linewidth}{!}{    \small
    \begin{tabular}{p{\linewidth}}
        \toprule
        \underline{\textbf{Prompt for Zero-Shot Setting}} \\
        \vspace{1mm}

\textsc{[[Instruction]]}

Please select the best answer to the following multiple-choice question based on the video. 

Only one option is the most accurate answer in relation to the question and the video.

What is the correct answer to this question 

\colorbox{mygray}{\{Question\}}  

Options:

\colorbox{mygray}{\{Options\}}  

\textsc{[[End Of Instruction]]}

\textsc{[[Output Format]]}

Format your answer as follows:

Please directly output the answer letter without thinking and explanation.

If the correct option letters (A, B, C, D... ) for the multiple-choice question is X,
give the final correct option number in the following format: "[[X]]"

\textsc{[[End Of Output Format]]} \\
    \bottomrule
    \end{tabular}}
    \label{appendix:evaluation_prompt_zero_shot}
    
\end{table*}

\begin{table*}[h]

\centering
    \caption{    
    Evaluation prompt for the Zero-Shot + CoT Setting.
    }

\resizebox{1.05\linewidth}{!}{    \small
    \begin{tabular}{p{\linewidth}}
        \toprule
        \underline{\textbf{Prompt for Zero-Shot + CoT Setting}} \\
        \vspace{1mm}


\textsc{[[Instruction]]}

Please select the best answer to the following multiple-choice question based on the video. 

Only one option is the most accurate answer in relation to the question and the video.

What is the correct answer to this Question:

\colorbox{mygray}{\{Question\}}  

Options:

\colorbox{mygray}{\{Options\}}  

Let's think step by step.

\textsc{[[End Of Instruction]]}

\textsc{[[Output Format]]}

Format your answer as follows:

Your thinking process.

If the correct option letters (A, B, C, D... ) for the multiple-choice question is X,
give the final correct option number in the following format: "[[X]]"

\textsc{[[End Of Output Format]]} \\
    \bottomrule
    \end{tabular}}
    \label{appendix:evaluation_prompt_cot}
    
\end{table*}

\section{Case Study}
\label{sec:case}
In this section, we present reasoning processes and results from selected models on the MMR-V benchmark. Through these case studies, we aim to better illustrate the current shortcomings of models in multimodal reasoning tasks and provide insights that may inspire future research and advancements in this area.

\begin{figure}[t]
    \centering
    \includegraphics[width=\linewidth]{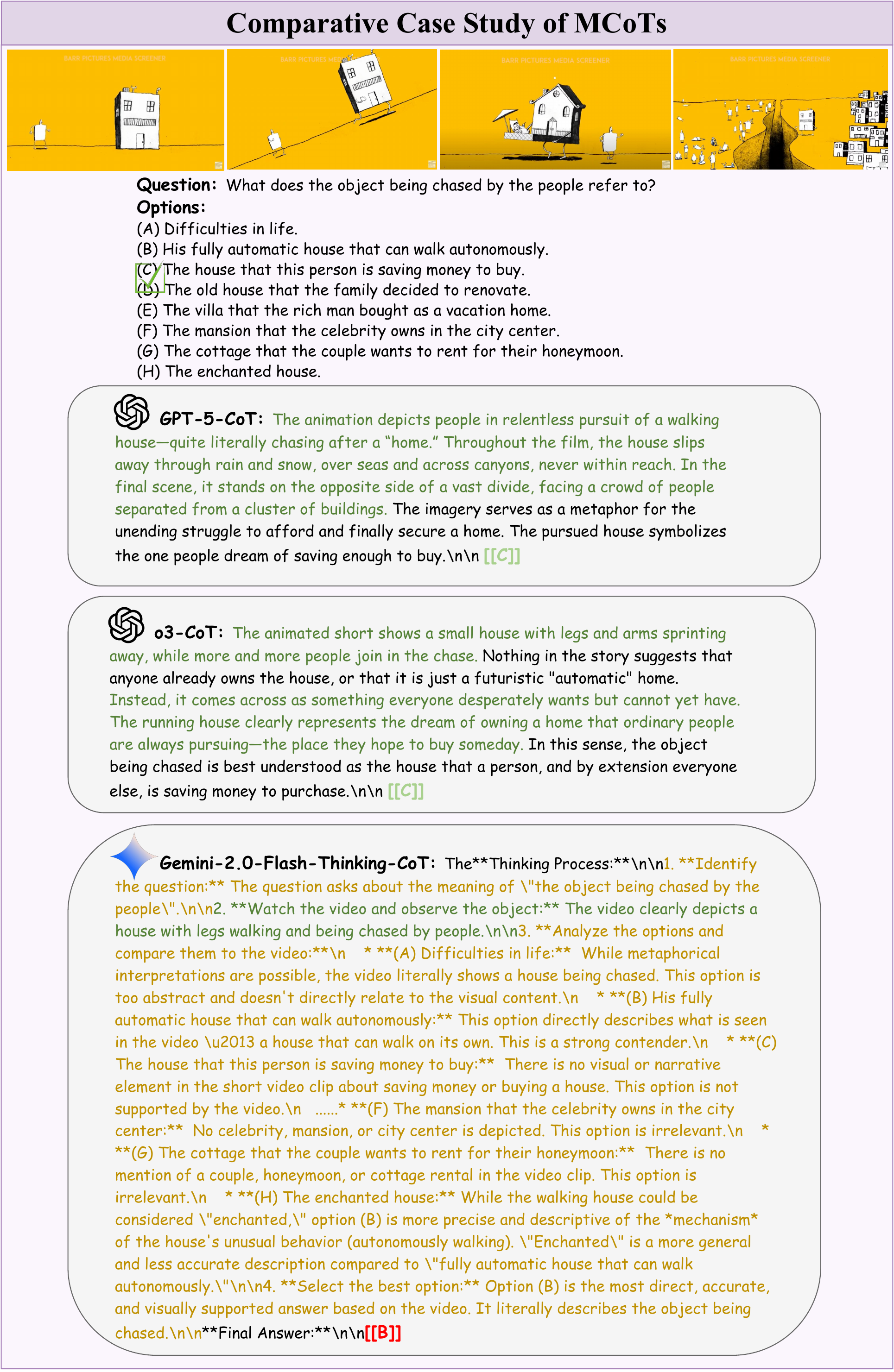}
    \caption{A comparison of CoTs from two models on the same task. Yellow and green indicate text and video analysis, respectively. As shown, GPT-5 and o3's reasoning paradigm demonstrates a deeper analysis of the video content.}
    \label{fig:good_vs_bad}
\vspace{-1em}
\end{figure}

Firstly, there is a comparison between a good CoT and a poor CoT in \ref{fig:good_vs_bad}. Yellow highlights indicate text-based analysis, while green highlights denote video-based analysis. As shown, the upper two CoTs engage in deep multimodal reasoning grounded in the video, ultimately reaching the correct answer. In contrast, the bottom CoT relies mainly on textual reasoning with only shallow video perception, leading to an incorrect outcome.

\begin{figure}[t]
    \centering
    \includegraphics[width=\linewidth]{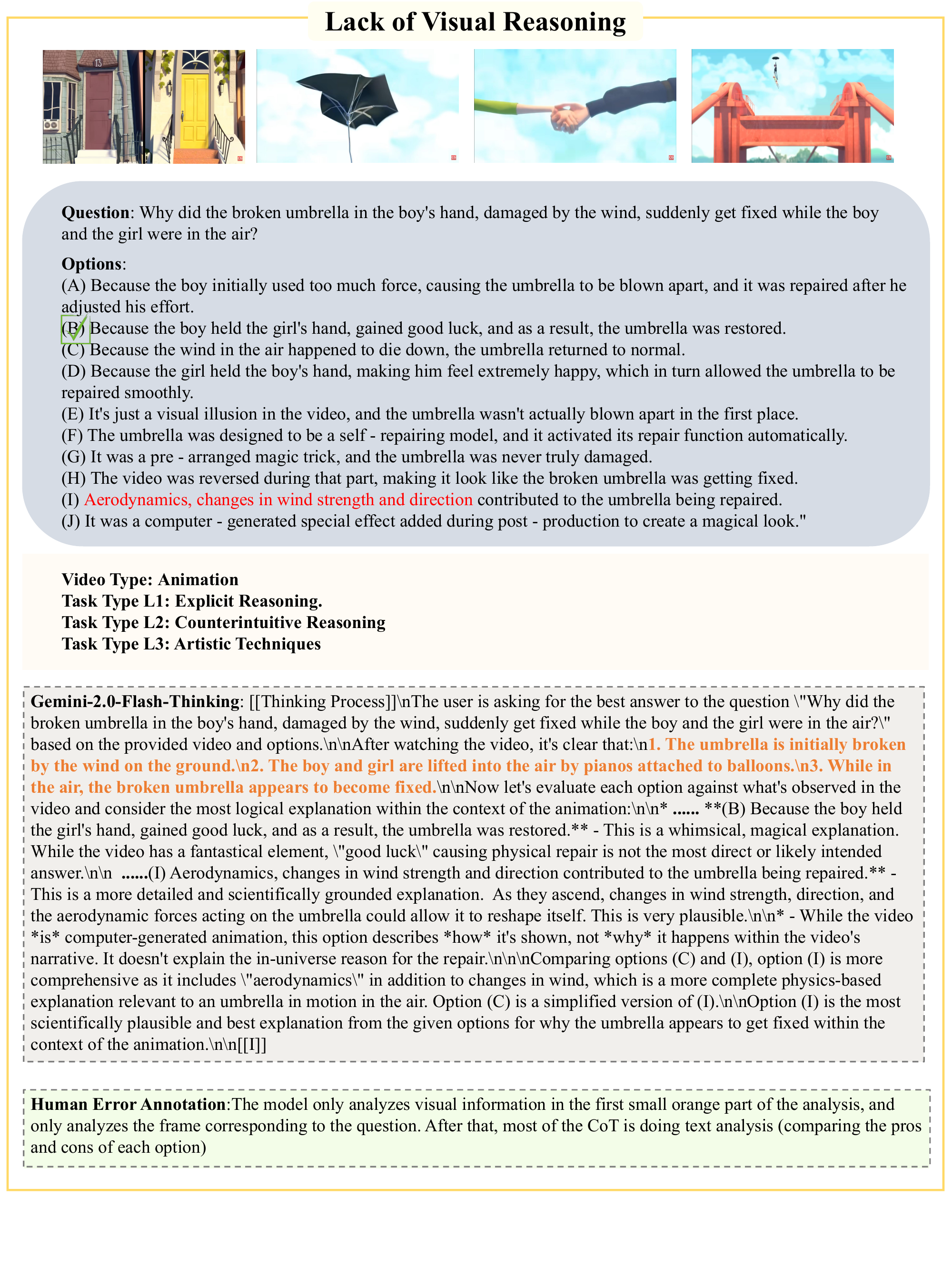}
    \caption{Error Case: Lack of Visual Reasoning.}
    \label{fig:visual reasoning error 1}
\vspace{-1em}
\end{figure}

\begin{figure}[t]
    \centering
    \includegraphics[width=\linewidth]{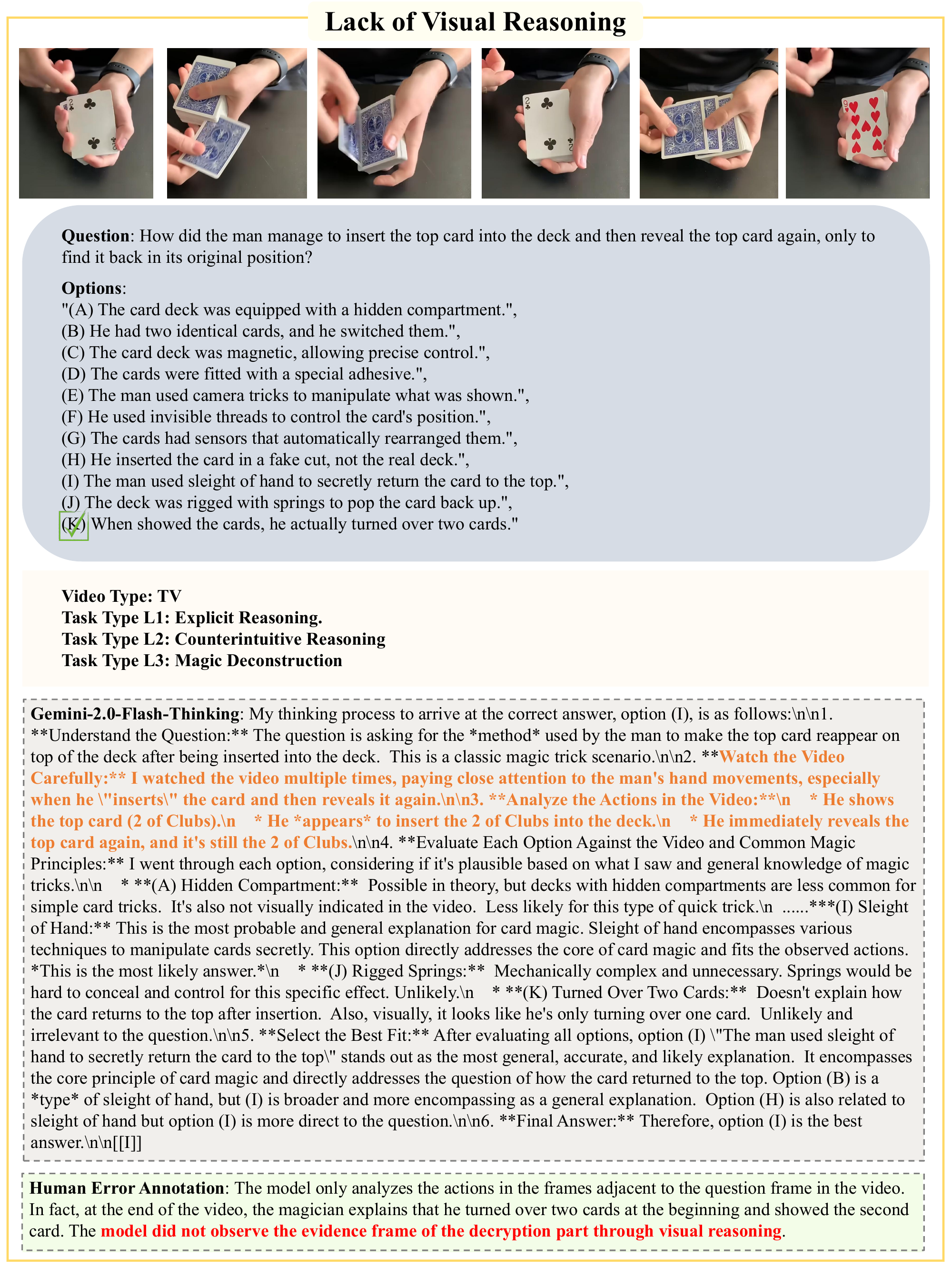}
    \caption{Error Case: Lack of Visual Reasoning.}
    \label{fig:visual reasoning error 2}
\vspace{-1em}
\end{figure}

\begin{figure}[t]
    \centering
    \includegraphics[width=\linewidth]{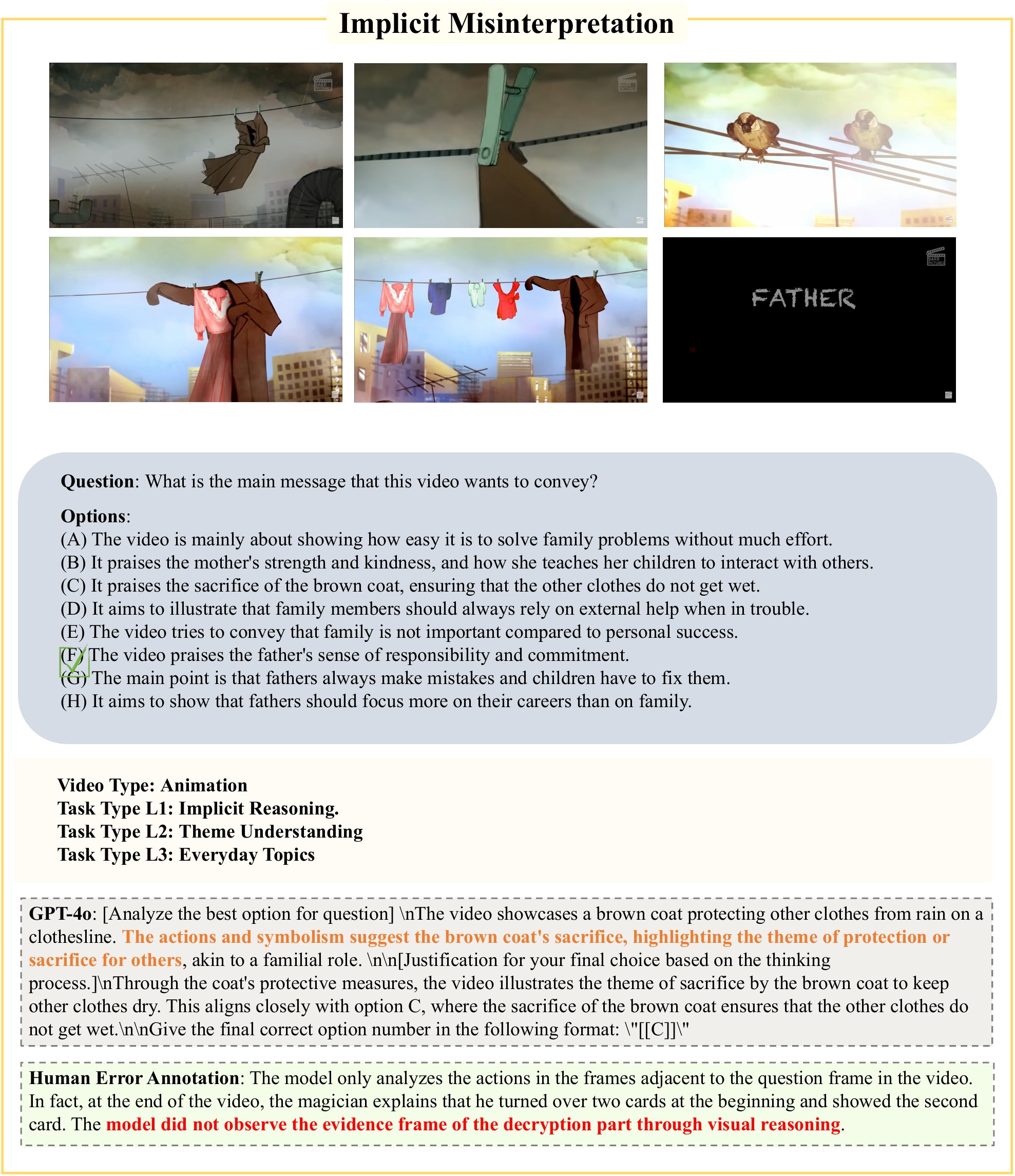}
    \caption{Error Case: Implicit Misinterpretation.}
    \label{fig:Implicit Misinterpretation error}
\vspace{-1em}
\end{figure}

\begin{figure}[t]
    \centering
    \includegraphics[width=\linewidth]{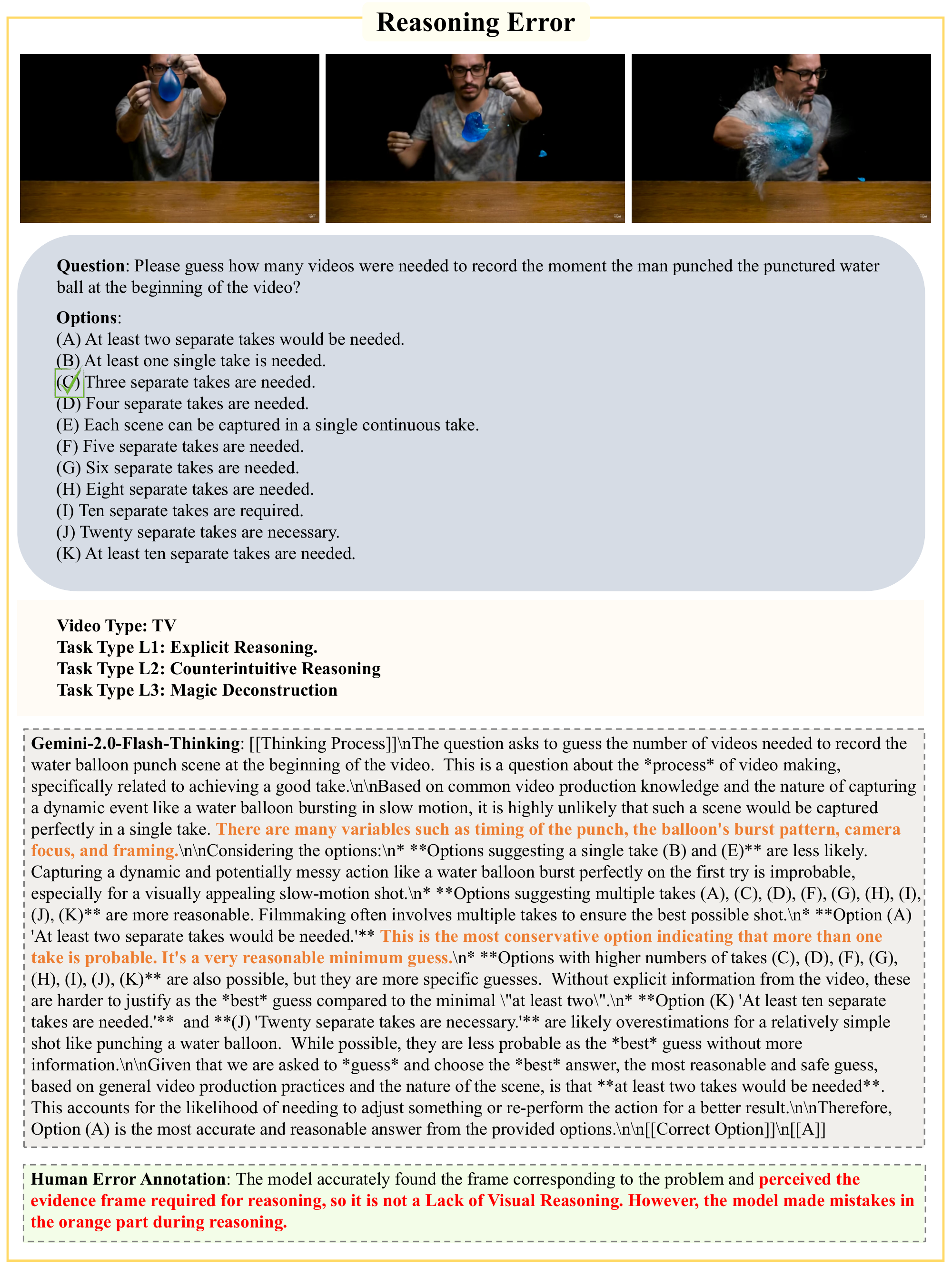}
    \caption{Error Case: Reasoning Error. The reasoning and analysis process of this example can refer to the \href{https://www.bilibili.com/video/BV1c3ZwY4EqE?spm_id_from=333.788.recommend_more_video.-1&vd_source=e2638f46408a99009fc4299e944cf139}{disassembly video}.}
    \label{fig:Reasoning Error}
\vspace{-1em}
\end{figure}


\begin{figure}[t]
    \centering
    \includegraphics[width=\linewidth]{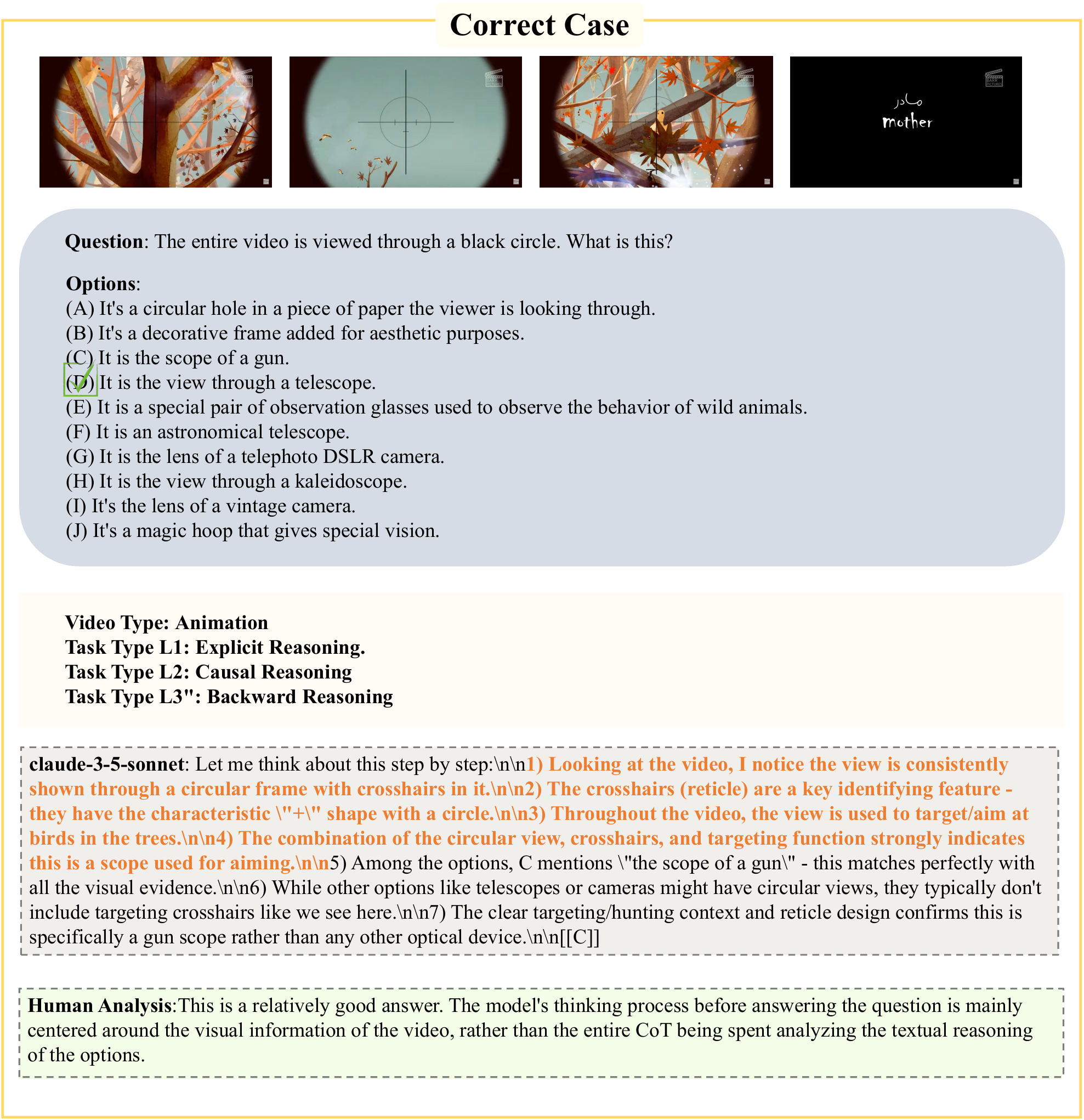}
    \caption{Correct Case.}
    \label{fig:good case 1}
\vspace{-1em}
\end{figure}

\end{document}